\documentclass[acmsmall,authorversion]{acmart}
\usepackage{amsmath}
\usepackage{mathtools}
\usepackage{amsthm}
\usepackage{graphicx}
\usepackage{booktabs}
\usepackage{array}
\usepackage{multirow}
\usepackage{color}
\usepackage{colortbl}
\usepackage{framed}
\usepackage{bm}
\usepackage{bbm}
\usepackage{xspace}
\usepackage{enumitem}
\usepackage{cite}
\usepackage{xparse}
\usepackage{xcolor}
\usepackage{algorithm}
\usepackage{lipsum}
\usepackage{listings}
\usepackage{url}
\usepackage{pifont}

\def \ie {\emph{i.e.}}

\def \etal {\emph{et al.}}

\newcommand{\tit}[1]{\smallbreak\noindent\textbf{#1.}}
\newcommand{\tinytit}[1]{\noindent\textbf{#1.}}

\newcommand{\rev}[1]{\textcolor{black}{#1}}
\newcommand{\revv}[1]{\textcolor{black}{#1}}

\AtBeginDocument{%
  }

\setcopyright{acmcopyright}
\copyrightyear{2024}
\acmYear{2024}

\acmJournal{TOMM}

\begin{document}

\title{Parents and Children:\\Distinguishing Multimodal DeepFakes from Natural Images}

\author{Roberto Amoroso}
\authornote{Both authors contributed equally to this research.}
\orcid{0000-0002-1033-2485}
\affiliation{%
  \institution{University of Modena and Reggio Emilia}
  \city{Modena}
  \country{Italy}
}
\email{roberto.amoroso@unimore.it}

\author{Davide Morelli}
\authornotemark[1]
\orcid{0000-0001-7918-6220}
\affiliation{%
  \institution{University of Modena and Reggio Emilia}
  \city{Modena}
  \country{Italy}
}
\affiliation{%
  \institution{University of Pisa}
  \city{Pisa}
  \country{Italy}
}
\email{davide.morelli@unimore.it}

\author{Marcella Cornia}
\orcid{0000-0001-9640-9385}
\affiliation{%
  \institution{University of Modena and Reggio Emilia}
  \city{Modena}
  \country{Italy}
}
\email{marcella.cornia@unimore.it}

\author{Lorenzo Baraldi}
\orcid{0000-0001-5125-4957}
\affiliation{%
  \institution{University of Modena and Reggio Emilia}
  \city{Modena}
  \country{Italy}
}
\email{lorenzo.baraldi@unimore.it}

\author{Alberto Del Bimbo}
\orcid{0000-0002-1052-8322}
\affiliation{%
  \institution{University of Florence}
  \city{Florence}
  \country{Italy}
}
\email{alberto.delbimbo@unifi.it}

\author{Rita Cucchiara}
\orcid{0000-0002-2239-283X}
\affiliation{%
  \institution{University of Modena and Reggio Emilia}
  \city{Modena}
  \country{Italy}
}
\affiliation{%
  \institution{IIT-CNR}
  \city{Pisa}
  \country{Italy}
}
\email{rita.cucchiara@unimore.it}

\renewcommand{\shortauthors}{R. Amoroso et al.}
\renewcommand{\shorttitle}{Parents and Children: Distinguishing Multimodal DeepFakes from Natural Images}

\begin{abstract}
Recent advancements in diffusion models have enabled the generation of realistic deepfakes from textual prompts in natural language. While these models have numerous benefits across various sectors, they have also raised concerns about the potential misuse of fake images and cast new pressures on fake image detection. In this work, we pioneer a systematic study on deepfake detection generated by state-of-the-art diffusion models. Firstly, we conduct a comprehensive analysis of the performance of contrastive and classification-based visual features, respectively extracted from CLIP-based models and ResNet or ViT-based architectures trained on image classification datasets. Our results demonstrate that fake images share common low-level cues, which render them easily recognizable. Further, we devise a multimodal setting wherein fake images are synthesized by different textual captions, which are used as seeds for a generator. Under this setting, we quantify the performance of fake detection strategies and introduce a contrastive-based disentangling method that lets us analyze the role of the semantics of textual descriptions and low-level perceptual cues. Finally, we release a new dataset, called COCOFake, containing about 1.2M images generated from the original COCO image-caption pairs using two recent text-to-image diffusion models, namely Stable Diffusion v1.4 and v2.0.
\end{abstract}

\ccsdesc[500]{Computing methodologies~Image representations}
\ccsdesc[500]{Computing methodologies~Matching}
\ccsdesc[300]{Computing methodologies~Computer vision tasks}

\keywords{multimodal deepfakes, vision-and-language, generative models}


\maketitle

\section{Introduction}
\label{sec:intro}
Machine-generated images have gained extensive popularity in the digital world due to the popularity of  GANs~\citep{goodfellow2014generative,mirza2014conditional,karras2019style,karras2020analyzing} and diffusion models~\citep{dhariwal2021diffusion,ramesh2022hierarchical,saharia2022photorealistic,rombach2022high}. While image generation tools can be employed for lawful goals, such as assisting content creators, generating simulated datasets, or enabling multimodal interactive applications, they have raised concerns regarding their potential for illegal and malicious purposes~\citep{brundage2018malicious,harris2018deepfakes,chesney2019deepfakes,cucchiara2024video}. These include the forgery of natural images, the generation of images in support of fake news, and the generation of NSFW contents~\citep{schramowski2023safe,poppi2023removing}. In this context, assessing the authenticity of images becomes a fundamental goal for security and for guaranteeing the trustworthiness of AI algorithms. 

\begin{figure}[t]
\centering
\includegraphics[width=.85\linewidth]{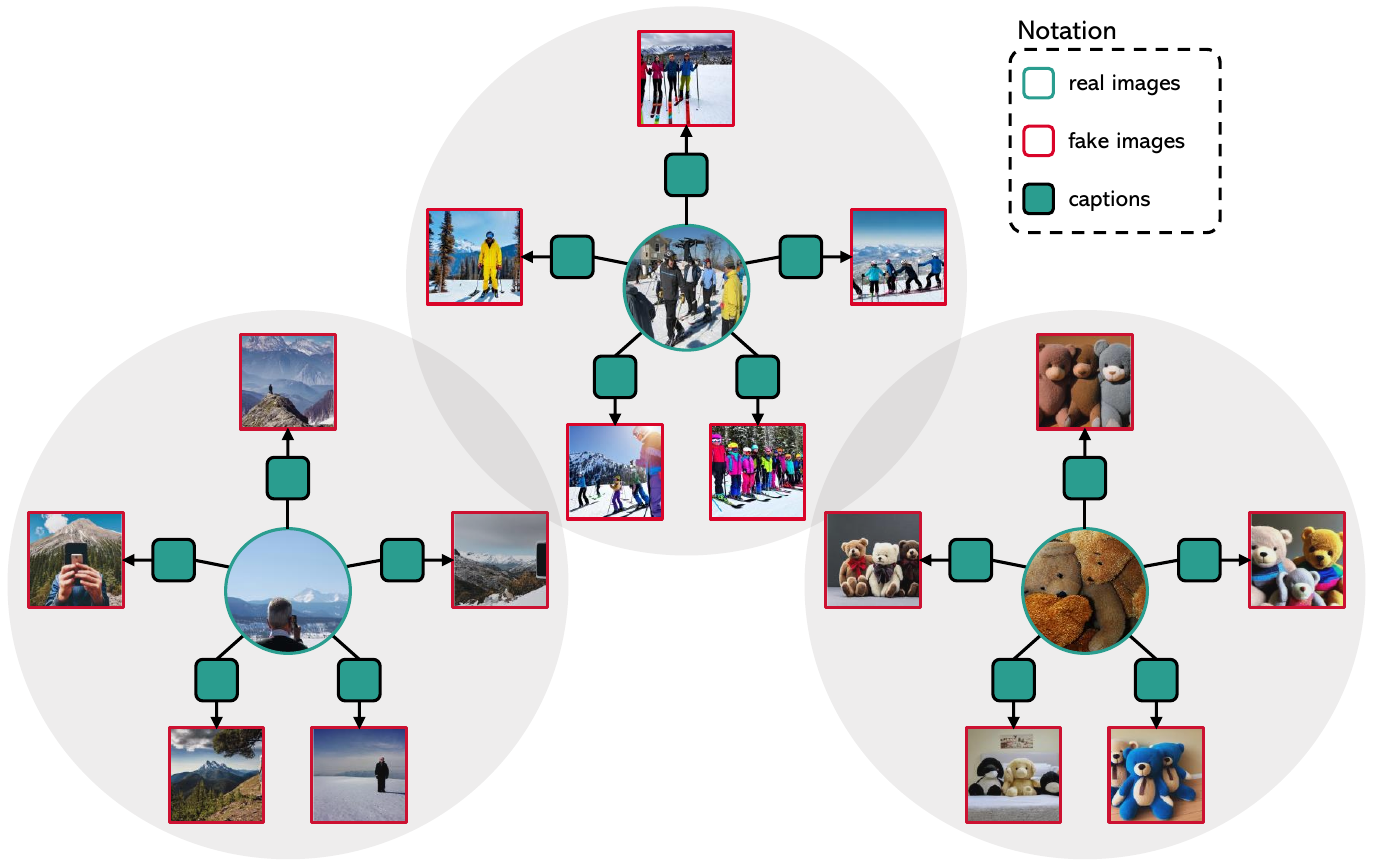}
\caption{Overview of our multimodal deepfakes detection setting, in which five subsets of the semantics contained in a given image are employed to generate as many fake images.}
\label{fig:firstpage}
\end{figure}

Most of the past approaches for deepfake detection have employed perceptual cues~\citep{zhang2019detecting,durall2020watch,frank2020leveraging}, including frequency analysis, the detection of artifacts, or pixel discontinuities. Furthermore, a significant portion of the early studies has focused exclusively on fake faces~\citep{li2018exposing,rossler2019faceforensics,li2020celeb}. Today's generators~\citep{ramesh2021zero,ding2021cogview,ramesh2022hierarchical,saharia2022photorealistic,rombach2022high,gafni2022make} are general-purpose, text-driven, and exhibit higher generation quality. If we look at images generated by Stable Diffusion~\citep{rombach2022high} (a few
examples are reported in Sec.~\ref{sec:dataset}), we might notice that some of them appear hyper-realistic and, thus, easily recognizable, while others contain semantic anomalies. However, most of them are realistically plausible.

In this paper, we aim at developing a systematic study on deepfake detection, in an era when generated content is becoming increasingly realistic and text-driven. We do this in a multimodal setting that enables us to examine deepfake detection from both a perceptual and a semantic perspective. Specifically, given an image, we consider different textual descriptions and fake images generated by using each of the descriptions as a prompt (Fig.~\ref{fig:firstpage}). In this manner, we build clusters sharing similar semantics, containing one real image and multiple fake images.
\rev{Under this setting, we first train a classifier to recognize deepfakes and investigate the effectiveness of different visual features extracted from both contrastive-based backbones like CLIP~\citep{radford2021learning} and classification-based ones such as ResNet~\citep{he2016deep} and ViT-based networks~\citep{dosovitskiy2020image} trained on ImageNet.} Surprisingly, we find out that high-level contrastive-based features learned on image and text pairs are very effective in discriminating between real and generated images. We hypothesize that low-level perceptual features also percolate into such descriptors, even though they are trained at a semantic level.

While these findings might be effective in defending us from current generators, we can expect that tomorrow's generators will increase their quality and become less detectable via low-level features. Thus, we devise a contrastive-based disentanglement strategy that enables to remove the contribution of low-level features. This approach establishes a more complex setting in which generated images cannot be distinguished at a perceptual level. Under this setting, we propose and discuss a general procedure for discriminating between fake and real images based on semantic information. 
\rev{To evaluate the effectiveness of the proposed method, we introduce a new dataset, namely COCOFake, which comprises approximately 1.2M images generated from the original COCO image-caption pairs using both Stable Diffusion v1.4 and v2.0 as text-to-image generative models.}

\smallbreak
\noindent \textbf{Contributions} In summary, the main contributions of this work are as follows:
\begin{itemize}[nosep]
    \item We develop a framework that utilizes machine-generated variants of natural images to investigate the detectability of diffusion model-generated images at the semantic and perceptual levels. By filtering the semantic content of natural images through natural language descriptions, we create a dataset of machine-generated images that can be used to investigate the performance of fake detection against modern diffusion models.
    \item We demonstrate that contrastive-based features can be effectively employed for fake detection against modern diffusion models, with high recognition rates.
    \item We propose a contrastive-based disentanglement approach to distinguish between low-level and semantic features in modern visual extractors. This allows us to distinguish between natural images and the generated ones using only semantic cues while neglecting the perceptual ones. This is important for the future development of more realistic generators.
    \item \rev{We generate and release the COCOFake dataset\footnote{\rev{The dataset can be downloaded at this link: \url{https://github.com/aimagelab/COCOFake}.}}, which contains over 1.2M fake images linked to natural images through captions. This dataset can be used to test and evaluate the performance of fake detection algorithms against diffusion model-generated images and assess their robustness in detecting fake images generated by different text-to-image generative models.}
\end{itemize}

\section{Related Work}
\label{sec:related}
\tinytit{General Deepfake Detection}
In recent years, with the growth and diffusion of generative models, several research efforts~\citep{verdoliva2020media,cozzolino2018forensictransfer} have been made to effectively detect synthetic images generated by GANs~\citep{goodfellow2014generative,mirza2014conditional,zhu2017unpaired,karras2019style,karras2020analyzing} and other deep learning-based architectures~\citep{kingma2018glow,vahdat2020nvae}. While initial works did not concentrate on the generalization capabilities of deepfake detectors~\citep{marra2018detection,rossler2019faceforensics}, subsequent approaches~\citep{chai2020makes,wang2020cnn,cozzolino2021towards,gragnaniello2021gan,girish2021towards,mandelli2022detecting} focused instead on the development of generic detectors that can be applied to different generators, thus avoiding the need to have a specific detector for each generative model. On the same line, different solutions~\citep{zhang2019detecting,durall2020watch,frank2020leveraging} proposed to detect deepfakes based on the spectrum of GAN-generated images. In fact, CNN-based generative models usually leave a distinguishable fingerprint over generated images, due to transposed convolutions~\citep{zhang2019detecting,durall2020watch}, up-sampling operations~\citep{frank2020leveraging,chandrasegaran2021closer}, and the spectral bias of convolution layers~\citep{dzanic2020fourier,khayatkhoei2022spatial}. Some works in similar directions also focused on associating fake images to the corresponding generator among several known GANs~\citep{yu2019attributing,joslin2020attributing} \revv{or extending deepfake detection to the video domain~\citep{cozzolino2021id,gu2021spatiotemporal,han2021fighting,yang2023avoid,han2023sigma}. In the latter case, deepfakes are usually generated by partially manipulating original videos with existing tools for face swapping and other sophisticated algorithms for audio manipulation. Research efforts in this domain have mainly been dedicated to improving deepfake detection performance with the integration of multiple modalities, such as spatial rich model filters~\citep{han2021fighting,luo2021generalizing} and audio traces~\citep{yang2023avoid,cheng2023voice} in both cases combined with RGB features.}

\tit{Detection of Deepfakes Generated with Diffusion Models}
While all aforementioned methods are tailored for detecting deepfakes generated by GANs or other visual forgery tools, a few works extended the analysis to deepfake images coming from diffusion models~\citep{dhariwal2021diffusion,ramesh2022hierarchical,nichol2021glide,saharia2022photorealistic,rombach2022high}. Among them,~\citet{wolter2022wavelet} proposed to detect fake images based on their wavelet-packet representations taking into account features from the pixel and frequency space.\rev{~\citet{ricker2022towards} evaluate the performance of state-of-the-art detectors and also tackle the frequency domain, analyzing different factors that influence the spectral properties of these images, discovering that GANs and diffusion models produce images with different characteristics that require adaptation of existing classifiers to ensure reliable detection}. Similarly,~\citet{corvi2022detection} introduced an analysis of the forensics traces left by common diffusion models and investigated whether deepfake detectors tailored for GANs can also distinguish images generated by diffusion models. Finally,~\citet{sha2022fake} analyzed and compared deepfakes generated by different text-to-image diffusion models, investigating the possibility of correctly attributing deepfake images to the diffusion model that generated them. Overall, these studies highlight the need for developing detection methods that can effectively detect deepfakes generated by various types of generative models, including diffusion models.

\tit{Datasets for Deepfake Detection}
The availability of large datasets has played a crucial role in the development of deepfake detection techniques. One of the most widely used datasets is FaceForensics++~\citep{rossler2019faceforensics}, which contains videos of real and fake faces generated using several generative models. The dataset provides both raw and manipulated videos with different compression rates and resolutions, allowing the evaluation of deepfake detection methods under different scenarios. Another popular dataset is Celeb-DF~\citep{li2020celeb}, which contains videos of celebrities manipulated using different techniques including GANs and face swapping. Celeb-DF also provides several levels of difficulty, ranging from low-quality to high-quality forgeries, making it suitable for evaluating both traditional and advanced deepfake detection methods. Other datasets have been proposed, such as \mbox{DeeperForensics-1.0}~\citep{jiang2020deeperforensics}, which contains manipulated videos generated using multiple GAN-based models, and DFDC~\citep{dolhansky2020deepfake}, composed of thousands of videos of real and fake faces.

Despite the availability of these datasets, there is still a need for more diverse and challenging datasets that reflect the increasing sophistication of deepfake generation methods. In particular, while current datasets mainly focus on faces, there is a lack of datasets for detecting deepfakes in other types of images, such as natural scenes. The proposed COCOFake dataset aims to address this limitation by providing a large-scale dataset of natural images and their corresponding synthetic images generated by diffusion models, along with natural language captions linking them. This allows for the evaluation of deepfake detection methods in a more complex and diverse context and also enables the development of methods that can identify semantic inconsistencies between natural and synthetic images.

\section{Proposed Method}
\label{sec:method}
\subsection{Notation and Preliminaries}
We propose a framework for studying and detecting multimodal generated fake images, which encompasses the identification and separation of their perceptual and semantic components. 
In the rest of the paper, we will employ the following notation: $I_R$ will indicate a natural (real) image, $C$ a textual description (\ie,~a caption), and $I_F$ will indicate a fake image produced by a generator. Under this setting, a \emph{parent} real image $I_R$ can be the seed for $N$ different \emph{children} fake images $I_{F,i}$ given a set of textual descriptions $\{C_{i}\}$ of $I_R$, with $i={1,...,N}$, by using each of the descriptions as prompt for the generator.

\tit{Semantic and Style Components of an Image}
The information content of an image can be credited to many factors. For simplicity, we assume that an image $I$, regardless of its authenticity, embodies two information contributions, namely a \emph{semantic component} $\mathcal{H}_{sem}(I)$ and a perceptual or \emph{style component} $\mathcal{H}_{sty}(I)$. The former represents the content that could be expressed in a textual sentence, while the latter describes the image appearance, encompassing elements such as colors, textures, brightness, and low-level visual cues. Given a real image $I_R$, we can therefore express its total information $\mathcal{H}$ as a function of its semantic and style components, as follows:
\begin{equation}
\mathcal{H}(I_R) = f(\mathcal{H}_{sem}(I_R), \mathcal{H}_{sty}(I_R)).
\end{equation}

However, when an image is described through a natural language sentence, only a portion of its semantics is actually conveyed inside the caption. In other words, natural language descriptions act as a filter for the semantic content of the image. Hence, we introduce $\Delta \mathcal{H}_{sem}(I, C)$ to represent the portion of semantic information described by a caption $C$. By analogy, we could say that the textual descriptions of an image act as DNA fragments that can be utilized to generate an offspring of images. 

\tit{Generating Offspring with Natural Language Utterances}
From an input image $I_R$ we can, therefore, extract $N$ semantic information subsets $\Delta \mathcal{H}_{sem}^i(I_R, \cdot)$ and feed them to a generator obtaining $N$ different fake images $I_{F,i}$, with $i = 1,...,N$. We define \textit{semantic cluster} the ensemble of the starting real image $I_R$ and the offspring of $N$ fake images $I_{F,i}$ generated from it.
For instance, given a real image dataset such as COCO~\citep{lin2014microsoft}, containing $K$ images, each represented by $N=5$ captions, we could create $K$ clusters of $N+1$ images with one parent and $N$ children.

\subsection{Learning to Discriminate Real and Fake images}
Once a dataset in the aforementioned form has been built, we first measure to what extent real and generated images can be discriminated independently from their membership to a semantic cluster. Instead of doing this by learning ad-hoc visual features, we investigate the usage of state-of-the-art pre-trained visual models. In other words, given a dataset containing both real and generated images, we develop a model that identifies real images by using visual features extracted with a pre-trained backbone. Regarding the generation of the images, in the following, we will employ Stable Diffusion~\citep{rombach2022high}, which is freely available and represents a state-of-the-art approach. Nevertheless, the approach could be easily extended to other generators. 

To evaluate the discriminative power of current pre-trained visual features, we model the discriminator as a two-class linear classifier, so that input visual features are only linearly projected before taking the final decision on their realism.
Formally, given a real image $I_R \in \mathbb{R}^{3 \times H \times W}$ and an image encoder $ E_I : \mathbb{R}^{3 \times H \times W} \to \mathbb{R}^{D}$, we extract a vectorial image feature $F_I$ as
\begin{equation}
F_I = E_I(I_R).
\end{equation}
The features $F_I$ are then fed into a linear layer $L: \mathbb{R}^D \to \mathbb{R}$, whose output is thresholded to classify between \textit{real} (\ie,~0) and \textit{fake} (\ie,~1) images. 
As it will be discussed in the experimental section, our findings indicate that this is (still) a relatively simple task even when employing a state-of-the-art generator. This is, most likely, due to the fact that fake images are slightly different in terms of low-level cues with respect to real images.

\subsection{Semantic Preservation Analysis}
\label{sec:disentangle}
As a second analysis, we investigate the preservation of semantic information across both real and generated fake images. To do so, we consider a multimodal embedding space, in which both images and texts can be projected. Specifically, we verify if, starting from a generated image, we can retrieve the particular caption used as prompt during its generation. In other words, we test if the subset of the real semantic information $\Delta \mathcal{H}_{sem}(I_F, C)$ associated with a caption $C$ is still recognizable in the visual features extracted from the generated image.
 
Formally, given a caption $C$ describing a real image $I_R$, and a textual encoder $E_T$, we tokenize and extract the textual features $F_T$ as:
\begin{equation}
F_T = E_T(C).
\end{equation}
For each visual feature of a given fake image $I_{F}$, we verify the ability to retrieve the corresponding textual feature used to create $I_F$ through the generator model.

As it will be shown in the experimental section, we find out that (a) the alteration of low-level cues induced by the generator does not affect the semantic contribution coming from the original image, and (b) the semantic contribution of the generator does not obfuscate the original semantic content.

\subsection{Disentangling Semantics and Style}
As the detection of fake images is likely promoted by the difference in low-level cues between generated and real images, we finally investigate a more challenging setting in which the style component induced by the generator is disentangled and removed. To do so, we learn a model which identifies the style component of the generator which is common to all generated images. We then measure whether, after eliminating such a component, the remaining semantic information is sufficient to discriminate between real and fake images. Noticeably, this corresponds to a more challenging setting where all the common low-level traits left by the generator are removed and not employed to perform deepfake classification. In other words, this also corresponds to recognizing fakes generated by an ``ideal'' generator that does not leave common low-level traits. 

\begin{figure}[t]
    \centering
    \includegraphics[width=0.95\linewidth]{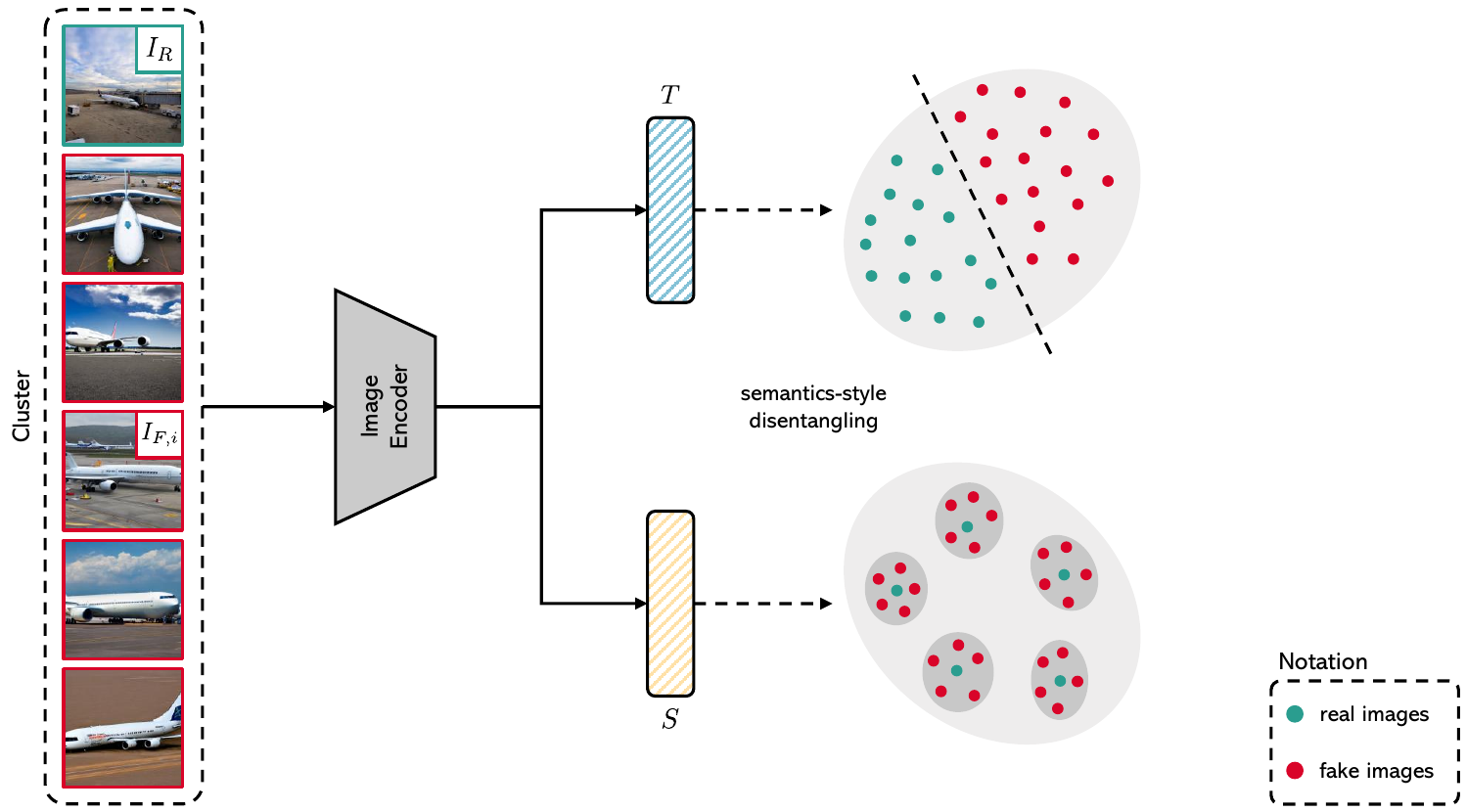}
    \caption{Schema of our approach for disentangling semantics and style for deepfake detection.}
    \label{fig:model}
\end{figure}

To perform this analysis, we propose a new contrastive-based learning model that can project images in a semantic space and in a style space (Fig.~\ref{fig:model}). For a good style-semantic disentanglement we expect that, in the style embedding latent space, the feature vectors of real images should be separated from features of fake images in a cluster-agnostic way, while in the semantic embedding latent space the cluster compactness should be preserved. Specifically, we train two separate linear projections $T$ and $S$, where $T$ focuses on style while $S$ on semantics.
For the $T$ layer we aim at increasing the distance between fake and real elements, regardless of their membership in a specific cluster. 
For the $S$ layer, instead, we want to create compact clusters of elements sharing the same semantic content, while increasing the distance among two fake elements or two real elements.

We express these requirements through two loss components $\mathcal{L}_{c}$ and $\mathcal{L}_{fr}$. The former attracts elements of the same cluster, while the latter attracts elements having the same label (\ie,~real and fake). From here, we can define the losses needed to train $T$ and $S$, respectively, as follows:
\begin{equation}
\begin{split}
&\mathcal{L}_{T} = \mathcal{L}_{fr}- \mathcal{L}_{c},  \\ 
&\mathcal{L}_{S} =  \mathcal{L}_{c}- \mathcal{L}_{fr}. 
\end{split}
\end{equation}
To implement both $\mathcal{L}_{c}$ and $\mathcal{L}_{fr}$, we leverage a Supervised Contrastive Loss~\citep{khosla2020supervised}, defined as follows:
\begin{equation}
\label{eq:supconloss}
\mathcal{L}_{SupCon} = \sum_{i \in I} \frac{-1}{\vert P(i) \vert} \sum_{p \in P(i)} \log{\frac{\exp(\mathcal{F}_i \mathcal{F}_p^\intercal/\tau)}{\sum\limits_{a \in A(i)} \exp(\mathcal{F}_i \mathcal{F}_a^\intercal/\tau)}},
\end{equation}
where $i \in I \equiv \{1,...,N+1\}$ represents the index of an arbitrary sample, $\mathcal{F}$ are $\ell_2$-normalized input features of a given image, $\tau$ is a temperature parameter, $A(i) \equiv I /\ \{i\}$. $P(i)$ is the set of indices of all items sharing the same label of $i$, and $\vert P(i) \vert$ is its cardinality.

\begin{figure}[t]
    \centering
    \resizebox{0.928\textwidth}{!}{
    \includegraphics[width=\textwidth]{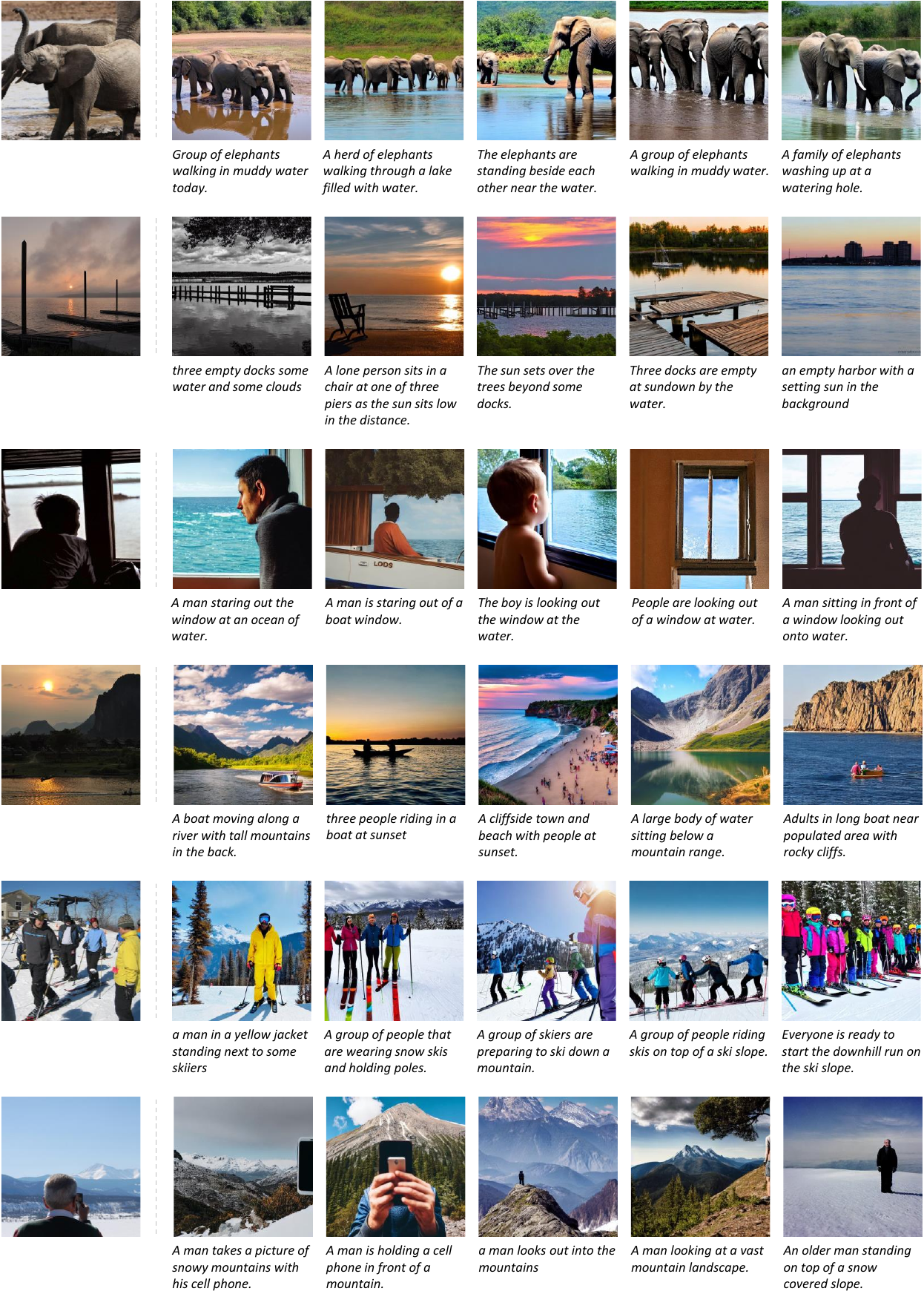}
    }
    \vspace{-0.2cm}
    \caption{\rev{Sample images from COCOFake. The leftmost column shows the original (real) image, while the remaining ones show fake images generated by Stable Diffusion v1.4 from each of the five COCO captions.}}
    \label{fig:dataset}
    \vspace{-0.2cm}
\end{figure}

Depending on the nature of the labels used in the training of the supervised contrastive loss, we can implement repulsive and attractive forces in the form of the loss components $\mathcal{L}_c$ and $\mathcal{L}_{fr}$.
In $\mathcal{L}_c$, in particular, we assign the same label to elements belonging to the same cluster, while in $\mathcal{L}_{fr}$ we assign the same label to all real samples, and the same label to all fake images. The objective of $\mathcal{L}_c$ is to attract elements of the same cluster, while $\mathcal{L}_{fr}$ pushes real and fake images. 

\begin{figure}[t]
    \centering
    \resizebox{0.928\textwidth}{!}{
    \includegraphics[width=\textwidth]{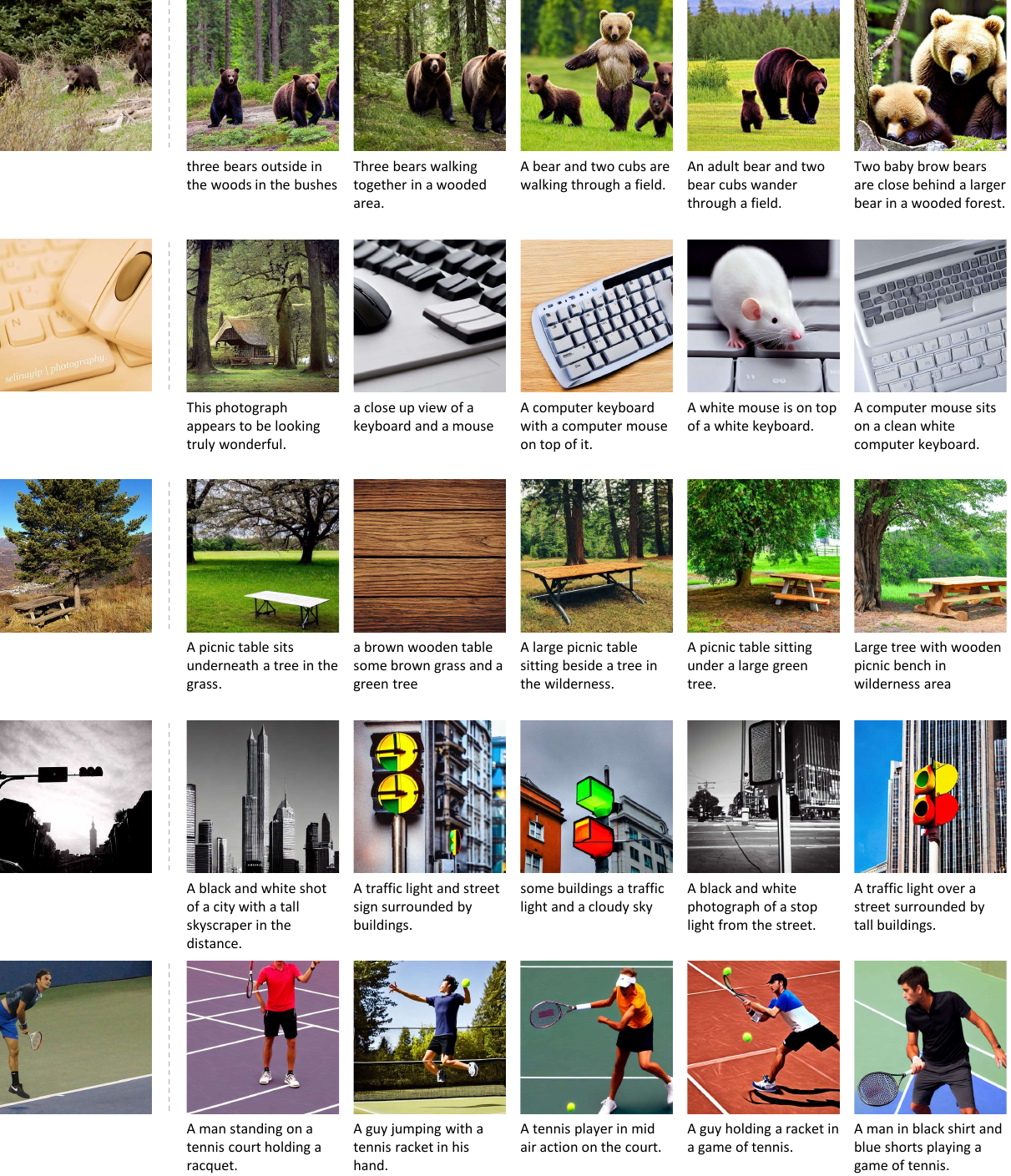}
    }
    \vspace{-0.2cm}
    \caption{\rev{Less realistic images from COCOFake. The leftmost column shows the original (real) image, while the remaining ones show fake images generated by Stable Diffusion v1.4 from each of the five COCO captions.}}
    \label{fig:dataset_bad_1}
    \vspace{-0.2cm}
\end{figure}

\subsection{The COCOFake Dataset for Multimodal Deepfake Recognition}\label{sec:dataset}
In literature, to the best of our knowledge, there are no multimodal datasets containing texts, real and fake images that are compatible with our multimodal setting. Thus, we generate and release the COCOFake dataset, an extension of COCO~\citep{lin2014microsoft}. Each real image in COCOFake is paired with five fake images that are conditionally generated based on each of the captions associated with the same image. We employ the Stable Diffusion model~\citep{rombach2022high} as our generator. \rev{Specifically, we create two different versions of our dataset, one based on Stable Diffusion v1.4\footnote{\rev{\url{https://huggingface.co/CompVis/stable-diffusion-v1-4}}} and the other based on Stable Diffusion v2.0\footnote{\rev{\url{https://huggingface.co/stabilityai/stable-diffusion-2-base}}}. Both text-to-image generators have been pre-trained on the English image-text pairs of the LAION-5B dataset~\citep{schuhmann2022laion} and finetuned on the LAION-Aesthetics subset\footnote{\url{https://laion.ai/blog/laion-aesthetics/}}. While Stable Diffusion v1.4 is based on the CLIP ViT-L/14 text encoder~\citep{radford2021learning}, the 2.0 version exploits the OpenCLIP ViT-H/14 one~\citep{radford2021learning}.} During image generation, we employ the safety checker module to reduce the probability of explicit images and disable the invisible watermarking of the outputs to prevent easy identification of the images as machine-generated. 

Overall, referring to the splits defined in~\citep{karpathy2015deep} and typically employed in image captioning literature~\citep{barraco2023little,sarto2023positive,caffagni2023synthcap}, the COCO dataset comprises 113,287 training images, 5,000 validation, and 5,000 test images. \rev{Preserving the same splits, COCOFake is composed of 679,722 training images, 30,000 validation, and 30,000 test images for each version of Stable Diffusion, thus comprising more than 1.2M generated images (\ie, around 600k for each version of Stable Diffusion).} Sample real-generated image clusters from the COCOFake dataset are shown in Fig.~\ref{fig:dataset}. For each example, we present the real image alongside the five fake images generated from each of the five captions from the original COCO dataset. As it can be seen, the generated images are generally coherent with the corresponding caption. However, in some cases, the generated images are overly realistic with brighter colors and a more professional photographic style than the real counterpart. This can be attributed to the dataset employed in the finetuning phase (\ie~the LAION-Aesthetics subset) of the Stable Diffusion model~\citep{rombach2022high}, used to generate fake images.
In Fig.~\ref{fig:dataset_bad_1} we report less realistic examples from the COCOFake dataset, again showing the original image and the five fake images with the corresponding captions. Failure cases include hallucinating the semantic content of the caption (first two rows), incorrect understanding of the caption (third row), abstract rendering of objects (traffic lights in the third row), and unrealistic rendering of human poses (last row). 

\rev{In our experiments, we evaluate deepfake detection performance under a standard setting in which we train the model on images generated by one Stable Diffusion version and test on images generated by the same model. Furthermore, to assess the robustness of our analysis, we also consider the generalization capabilities to images generated by a text-to-image diffusion model different from the one used during training.} Under this setting, we compare the performance of our method on images generated by different versions of Stable Diffusion, providing insights into the impact of the generative model on the deepfake detection performance.

\section{Experimental Evaluation}
\label{sec:experiments}
\subsection{Implementation Details}

\tinytit{Image Encoders} We test two families of backbones: the first are trained for classification on ImageNet~\citep{russakovsky2015imagenet}, while the second are trained on a cross-modal setting on large-scale datasets using contrastive-based loss functions. Due to the nature of the task these networks were trained for, only the latter family provides also text encoders $E_T$. Specifically, we employ a ResNet~\citep{he2016deep} model with 48 convolutional layers and a Vision Transformer (ViT)~\citep{dosovitskiy2020image} architecture in its B/32 configuration. The ViT encoder takes as input squared patches extracted from the input image and consists of a sequence of multi-head self-attention layers~\citep{vaswani2017attention}. Both these architectures are trained on the ImageNet dataset~\citep{russakovsky2015imagenet} that contains around 1.3M images.

As cross-modal architectures, we use two models coming from CLIP~\citep{radford2021learning}. In particular, we employ CLIP RN50 and CLIP ViT-B/32 models, both pre-trained on the OpenAI WebImageText (WIT) dataset, composed of 400 million image-text pairs collected from the web. Moreover, we employ the open source implementation of CLIP (\ie, OpenCLIP~\citep{wortsman2022robust}), trained with a post-ensemble method for improving robustness to out-of-distribution samples. In our experiments, we consider two versions of the OpenCLIP ViT-B/32 model: one trained on the LAION-400M dataset~\citep{schuhmann2021laion} that contains 400 million CLIP-filtered image-text pairs crawled from the web and the other trained on the larger LAION-2B composed of 2 billion image-text pairs~\citep{schuhmann2022laion}.

\tit{Linear Probing Details} In our experiments, we also conduct linear probes. In this case, we follow the approach of~\citep{radford2021learning} and employ the features extracted from the backbones to train a logistic regression model with $\ell_2$ penalty and LBFGS solver~\citep{byrd1995limited, zhu1997algorithm}. To balance the training samples, we employ one randomly extracted fake image for each cluster.

\tit{Disentanglement Architecture and Training Details}
When disentangling semantics and styles, we train the two linear layers $S$ and $T$, which perform a linear projection to the same dimensionality of the backbone visual features. To train these layers, we employ AdamW~\citep{loshchilov2018decoupled} as optimizer with $\beta_1 = 0.9$ and $\beta_2 = 0.999$. We use a batch size of 1,024 and a learning rate of 0.001, training all models for 25 epochs. 

\subsection{Metrics}
\rev{To assess the performance of our proposed methodology and evaluate spatial relationships between elements in the embedding spaces, we employ seven different metrics.} These aim to quantify the capability to discriminate between real and fake images and to quantify disentanglement.
 
\tit{Min and Max Intra-Cluster Distance Accuracy}
These two metrics are employed to evaluate the relative spatial positions of the elements inside a cluster. In particular, for each cluster, we measure the distances between the real image and each of the fake images belonging to the cluster. We then check how many times the real image is the item having the minimum or maximum distance with respect to all the others in the cluster. In other words, for each cluster, the min distance accuracy scores if the real image feature is on average the nearest to all the fake image features, while the max distance accuracy scores if it is the most distant one.

\tit{Overall and Full Cluster Accuracy}
These two metrics measure the real/fake classification accuracy both over the entire dataset and inside each cluster. The former metric is cluster-independent and is computed using all the elements of a dataset split (\ie,~validation, test). The latter, instead, is a cluster-based metric that scores if all elements of a cluster are correctly classified as real or fake, and the metric is then averaged across all clusters.

\tit{\rev{Overall AUC}} \rev{As reported in previous deepfake detection literature~\citep{mandelli2022detecting,corvi2022detection}, this metric is used along with accuracy to evaluate how well a deepfake detection model can distinguish between real and fake images. In our setting, it is computed using all the elements of the validation or test set of our dataset.}

\tit{Exact Pair and Intra-Cluster Retrieval}
These metrics are used to evaluate the goodness of the retrieval task (see Sec.~\ref{sec:disentangle}), in which given a generated image we seek to retrieve its parent caption. The former metric is a recall@k computed considering as ground-truth, for each fake image, the caption used for generating it. The latter, instead, is a recall@k that measures for a given fake image if the retrieved caption matches one of the five captions of the cluster the image belongs to.

\begin{table}[t]
\caption{Minimum and maximum distance accuracy on validation and test sets of COCOFake, using different visual backbones. \rev{Results are reported using images generated by both Stable Diffusion v1.4 and v2.0.}}
\label{tab:centroids}
\centering
\small
\setlength{\tabcolsep}{.35em}
\resizebox{\linewidth}{!}{
\begin{tabular}{lcc cc c cc c cc c cc}
\toprule
& & & \multicolumn{2}{c}{\textbf{Validation Set}} & & \multicolumn{2}{c}{\textbf{Test Set}} & & \multicolumn{2}{c}{\textbf{Validation Set}} & & \multicolumn{2}{c}{\textbf{Test Set}} \\
& & & \multicolumn{2}{c}{\textbf{(SD v1.4)}} & & \multicolumn{2}{c}{\textbf{(SD v1.4)}} & & \multicolumn{2}{c}{\textbf{(SD v2.0)}} & & \multicolumn{2}{c}{\textbf{(SD v2.0)}} \\
\cmidrule{4-5} \cmidrule{7-8} \cmidrule{10-11} \cmidrule{13-14}
& & & Min Dist. & Max Dist. & & Min Dist. & Max Dist. & & Min Dist. & Max Dist. & & Min Dist. & Max Dist. \\
\textbf{Backbone} & \textbf{Dataset} & & Accuracy & Accuracy  & & Accuracy & Accuracy & & Accuracy & Accuracy & & Accuracy & Accuracy \\
\midrule
RN50 & ImageNet & & 8.50 & 23.58 & & 8.82 & 24.82 & & 5.98 & 29.62 & & 6.62 & 30.16 \\
ViT-B/32 & ImageNet & & 6.84 & 23.12 & & 6.88 & 23.88 & & 5.12 & 29.18 & & 4.92 & 30.00 \\
\midrule
CLIP RN50 & OpenAI WIT & & 3.72 & 38.48 & & 3.60 & 41.24 & & 2.40 & 46.72 & & 2.20 & 48.28 \\
CLIP ViT-B/32 & OpenAI WIT & & 3.30 & 38.88 & & 3.24 & 40.10 & & 2.92 & 42.08 & & 2.98 & 44.18 \\
\midrule
OpenCLIP ViT-B/32 & LAION-400M & & 5.28 & 31.94 & & 5.00 & 32.02 & & 4.58 & 34.06 & & 4.62 & 36.02 \\
OpenCLIP ViT-B/32 & LAION-2B & & 1.40 & 42.80 & & 1.72 & 44.00 & & 1.88 & 42.64 & & 1.78 & 43.80 \\
\bottomrule
\end{tabular}
 }
 \vspace{-0.1cm}
\end{table}

\subsection{Performance of Visual Features}\label{sec:linear_probing}
\tinytit{Unsupervised Classification}
We start by assessing the capabilities of existing image features to discriminate between real and generated images, in an unsupervised setting. We employ the min and max distance accuracy metrics defined above and check the presence of spatial relationships between real and generated images inside each cluster.

Results are reported in Table~\ref{tab:centroids} on the test and validation sets of both Stable Diffusion v1.4 and v2.0. We employ six different visual backbones, namely two ResNet-50 pre-trained on ImageNet and OpenAI WIT and four ViT-B/32 pre-trained on ImageNet, OpenAI WIT, LAION-400M, and LAION-2B. As it can be seen, according to the features extracted from the aforementioned backbones, the real image of each cluster tends to be the one with maximum distance with respect to all the other elements. This suggests that these features are discriminative for the task of deepfake classification and that they percolate low-level features that allow for distinction between real and generated items inside of each semantic cluster. Noticeably, the maximum distance accuracy increases when considering backbones trained on multimodal datasets compared to backbones trained on classification, suggesting that image-text matching promotes the percolation of perceptual features.

\rev{Comparing the results when using fake images generated by the two considered Stable Diffusion versions, it can be noticed that Stable Diffusion v2.0 exhibits an improvement over v1.4 as evidenced by an increase in the maximum distance metric and a decrease in the minimum distance metric.} This suggests that the features extracted from v2.0 are better separable and hence the generated images are more easily detected.

\begin{table}[t]
\caption{Overall and full cluster accuracy results on the validation and test sets, using linear probing and features of different backbones trained on the COCOFake training set. \rev{Results are reported using images generated by both Stable Diffusion v1.4 and v2.0.}}
\label{tab:linprob}
\centering
\small
\setlength{\tabcolsep}{.3em}
\resizebox{\linewidth}{!}{
\begin{tabular}{lcc cc c cc c cc c cc}
\toprule
& & & \multicolumn{2}{c}{\textbf{Validation Set}} & & \multicolumn{2}{c}{\textbf{Test Set}} & & \multicolumn{2}{c}{\textbf{Validation Set}} & & \multicolumn{2}{c}{\textbf{Test Set}} \\
& & & \multicolumn{2}{c}{\textbf{(SD v1.4 $\rightarrow$ SD v1.4)}} & & \multicolumn{2}{c}{\textbf{(SD v1.4 $\rightarrow$ SD v1.4)}} & & \multicolumn{2}{c}{\textbf{(SD v1.4 $\rightarrow$ SD v2.0)}} & & \multicolumn{2}{c}{\textbf{(SD v1.4 $\rightarrow$ SD v2.0)}} \\
\cmidrule{4-5} \cmidrule{7-8} \cmidrule{10-11} \cmidrule{13-14}
& & & Overall & Full Cluster & & Overall & Full Cluster & & Overall & Full Cluster & & Overall & Full Cluster \\
\textbf{Backbone} & \textbf{Dataset} & & Accuracy & Accuracy & & Accuracy & Accuracy & & Accuracy & Accuracy & & Accuracy & Accuracy \\
\midrule
RN50 & ImageNet & & 90.31 & 57.56 & & 90.62 & 57.94 & & 81.71 & 34.94 & & 82.31 & 35.84 \\                   
ViT-B/32 & ImageNet & & 87.64 & 47.62 & & 87.16 & 47.32 & & 76.71 & 24.68 & & 77.31 & 26.92 \\               
\midrule
CLIP RN50 & OpenAI WIT & & 99.07 & 94.60 & & 99.17 & 95.30 & & 93.54 & 69.08 & & 93.74 & 69.64 \\            
CLIP ViT-B/32 & OpenAI WIT & & 99.11 & 94.84 & & 98.97 & 94.24 & & 94.41 & 72.30 & & 94.72 & 73.62 \\        
\midrule
OpenCLIP ViT-B/32 & LAION-400M & & 97.88 & 88.18 & & 97.83 & 87.80 & & 83.30 & 38.48 & & 84.32 & 40.74 \\    
OpenCLIP ViT-B/32 & LAION-2B & & 99.68 & 98.01 & & 99.64 & 97.84 & & 98.88 & 93.68 & & 98.96 & 94.08 \\      
\midrule
\midrule
& & & \multicolumn{2}{c}{\rev{\textbf{Validation Set}}} & & \multicolumn{2}{c}{\rev{\textbf{Test Set}}} & & \multicolumn{2}{c}{\rev{\textbf{Validation Set}}} & & \multicolumn{2}{c}{\rev{\textbf{Test Set}}} \\
& & & \multicolumn{2}{c}{\rev{\textbf{(SD v2.0 $\rightarrow$ SD v2.0)}}} & & \multicolumn{2}{c}{\rev{\textbf{(SD v2.0 $\rightarrow$ SD v2.0)}}} & & \multicolumn{2}{c}{\rev{\textbf{(SD v2.0 $\rightarrow$ SD v1.4)}}} & & \multicolumn{2}{c}{\rev{\textbf{(SD v2.0 $\rightarrow$ SD v1.4)}}} \\
\cmidrule{4-5} \cmidrule{7-8} \cmidrule{10-11} \cmidrule{13-14}
& & & \rev{Overall} & \rev{Full Cluster} & & \rev{Overall} & \rev{Full Cluster} & & \rev{Overall} & \rev{Full Cluster} & & \rev{Overall} & \rev{Full Cluster} \\
\rev{\textbf{Backbone}} & \rev{\textbf{Dataset}} & & \rev{Accuracy} & \rev{Accuracy} & & \rev{Accuracy} & \rev{Accuracy} & & \rev{Accuracy} & \rev{Accuracy} & & \rev{Accuracy} & \rev{Accuracy} \\
\midrule
\rev{RN50} & \rev{ImageNet} & & \rev{91.07} & \rev{59.84} & & \rev{91.45} & \rev{61.44} & & \rev{91.08} & \rev{60.44} & & \rev{91.33} & \rev{60.76} \\
\rev{ViT-B/32} & \rev{ImageNet} & & \rev{85.55} & \rev{42.92} & & \rev{86.12} & \rev{44.90} & & \rev{84.89} & \rev{41.50} & & \rev{84.49} & \rev{39.60} \\
\midrule
\rev{CLIP RN50} & \rev{OpenAI WIT} & & \rev{98.67} & \rev{92.56} & & \rev{98.68} & \rev{92.60} & & \rev{98.57} & \rev{91.94} & & \rev{98.66} & \rev{92.48} \\
\rev{CLIP ViT-B/32} & \rev{OpenAI WIT} & & \rev{98.56} & \rev{92.04} & & \rev{98.48} & \rev{91.48} & & \rev{98.58} & \rev{92.02} & & \rev{98.48} & \rev{91.76} \\
\midrule
\rev{OpenCLIP ViT-B/32} & \rev{LAION-400M} & & \rev{95.03} & \rev{74.70} & & \rev{95.57} & \rev{77.42} & & \rev{97.40} & \rev{85.62} & & \rev{97.29} & \rev{84.88} \\
\rev{OpenCLIP ViT-B/32} & \rev{LAION-2B} & & \rev{99.52} & \rev{97.16} & & \rev{99.59} & \rev{97.54} & & \rev{99.47} & \rev{96.80} & & \rev{99.41} & \rev{96.56} \\
\bottomrule
\end{tabular}
}
\vspace{-0.1cm}
\end{table}

\tit{Linear Probing}
Following the approach popularized by~\citep{radford2021learning}, we train a linear projection through logistic regression on top of the features extracted from the aforementioned backbones. \rev{We perform this experiment by training on both Stable Diffusion v1.4 and v2.0 images, and testing either on the validation and test sets containing images generated by the same Stable Diffusion version used during training or on the validation and test sets containing images generated by the Stable Diffusion model not used to train the linear projection.} 

\rev{Results are reported in Table~\ref{tab:linprob} in terms of overall accuracy and full cluster accuracy. As it can be seen, all the selected visual features exhibit a significant capability in linearly discriminating real and fake images, on the validation and test sets of the COCOFake dataset when considering both Stable Diffusion v1.4 and v2.0. In continuity with the previous experiment, we observe that contrastive-based visual backbones showcase significantly higher accuracy levels, up to 98.01\% and 97.16\% of full cluster accuracy respectively on the validation set with Stable Diffusion v1.4 and v2.0 images, and up to 99.68\% and 99.52\% overall accuracy on the same split.} This further confirms the observation that contrastive-based backbones extract and project into their embedding space, low-level and perceptual features that allow discriminating current deepfakes.
\rev{To assess the robustness of the method, we further test the trained classifiers on the data generated by the Stable Diffusion model not used during training (\ie, Stable Diffusion v2.0 for the linear projection trained on the 1.4 version, and Stable Diffusion v1.4 for the linear projection trained on the 2.0 version). As it can be observed in the right part of Table~\ref{tab:linprob}, the trained classifier performs comparably also in this setting with an overall accuracy close to or greater than 99\% in all cases. In particular, training on Stable Diffusion v2.0 images generalizes slightly better on images generated by Stable Diffusion v1.4 than the opposite direction with 99.47\% and 96.80\% of overall and full cluster validation accuracy compared to 98.88\% and 93.68\% obtained when testing the linear projection trained on Stable Diffusion v1.4 images on the validation set with images generated by the 2.0 version.}
Overall, these experiments show that the pre-trained visual backbones exhibit high discrimination power when identifying deepfakes.

\begin{figure}[t]
\centering
\footnotesize
\setlength{\tabcolsep}{.35em}
\resizebox{\linewidth}{!}{
\begin{tabular}{cc}
\textbf{Real Images} & \textbf{Fake Images} \\
\includegraphics[height=0.33\linewidth]{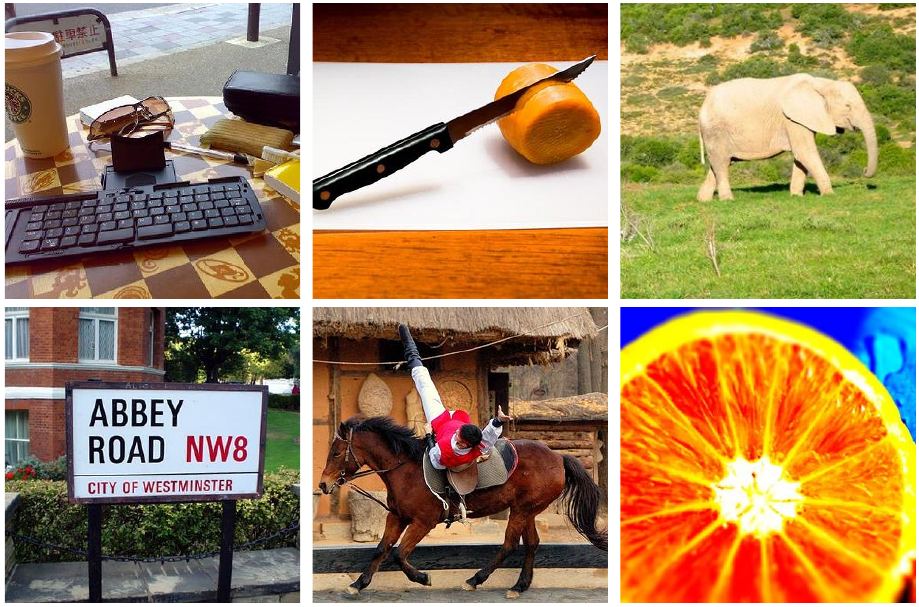} &
\includegraphics[height=0.33\linewidth]{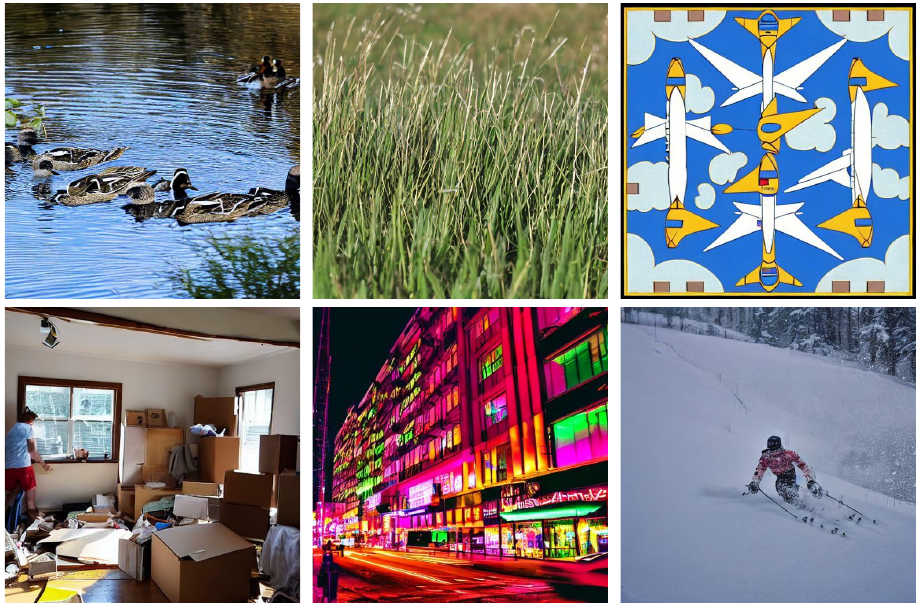} \\
\end{tabular}
}
\vspace{-0.2cm}
\caption{Sample misclassification errors on both real (left) and fake (right) images, using OpenCLIP ViT-B/32 trained on LAION-2B as the visual encoder.}
\label{fig:failures}
\vspace{-0.1cm}
\end{figure}

In light of the high accuracy levels of the aforementioned experiment, in Fig.~\ref{fig:failures} we report sample misclassified images. It can be noted, in particular, that fake images incorrectly classified as authentic (right side of the figure) depict close-ups and artistic drawings, whose authenticity is visually harder to guarantee.

\tit{Semantic Preservation}
We then conduct the retrieval-based analysis anticipated in Sec.~\ref{sec:disentangle}, in which we look for the original caption used to generate a particular image inside of a multimodal embedding space. The objective of this experiment is to assess whether the semantic information contained in the caption is preserved after the generation and to what extent the generation process alters semantic features. 

\begin{table}[t]
\caption{Exact pair and intra-cluster retrieval results. \rev{Results are reported using images generated by both Stable Diffusion v1.4 and v2.0.}}
\label{tab:retrieval}
\centering
\small
\setlength{\tabcolsep}{.32em}
\resizebox{\linewidth}{!}{
\begin{tabular}{lcc ccccccc c ccccccc}
\toprule
& & & \multicolumn{7}{c}{\textbf{Validation Set (SD v1.4)}} & & \multicolumn{7}{c}{\textbf{Test Set (SD v1.4)}} \\
\cmidrule{4-10} \cmidrule{12-18}
& & & \multicolumn{3}{c}{\textbf{Exact Pair}} & & \multicolumn{3}{c}{\textbf{Intra-Cluster}} & & \multicolumn{3}{c}{\textbf{Exact Pair}} & & \multicolumn{3}{c}{\textbf{Intra-Cluster}} \\
\cmidrule{4-6} \cmidrule{8-10} \cmidrule{12-14} \cmidrule{16-18}
\textbf{Backbone} & \textbf{Dataset} & & R@1 & R@3 & R@5 & & R@1 & R@3 & R@5 & & R@1 & R@3 & R@5 & & R@1 & R@3 & R@5 \\
\midrule
CLIP RN50 & OpenAI WIT & & 31.33 & 49.05 & 56.93 & & 41.91 & 58.46 & 66.01 & & 30.98 & 48.38 & 56.42 & & 42.09 & 58.35 & 65.93 \\
CLIP ViT-B/32 & OpenAI WIT & & 32.12 & 50.43 & 58.36 & & 43.34 & 60.15 & 67.42 & & 31.96 & 49.67 & 57.51 & & 43.24 & 59.3 & 66.78 \\
OpenCLIP ViT-B/32 & LAION-400M & & 36.48 & 55.36 & 63.28 & & 47.17 & 63.62 & 70.73 & & 35.53 & 54.49 & 62.56 & & 46.72 & 62.92 & 70.22 \\
OpenCLIP ViT-B/32 & LAION-2B & & 40.34 & 59.44 & 67.18 & & 50.78 & 66.64 & 73.58 & & 39.57 & 58.78 & 66.18 & & 50.46 & 66.34 & 73.03 \\
\midrule
\midrule
& & & \multicolumn{7}{c}{\rev{\textbf{Validation Set (SD v2.0)}}} & & \multicolumn{7}{c}{\rev{\textbf{Test Set (SD v2.0)}}} \\
\cmidrule{4-10} \cmidrule{12-18}
& & & \multicolumn{3}{c}{\rev{\textbf{Exact Pair}}} & & \multicolumn{3}{c}{\rev{\textbf{Intra-Cluster}}} & & \multicolumn{3}{c}{\rev{\textbf{Exact Pair}}} & & \multicolumn{3}{c}{\rev{\textbf{Intra-Cluster}}} \\
\cmidrule{4-6} \cmidrule{8-10} \cmidrule{12-14} \cmidrule{16-18}
\rev{\textbf{Backbone}} & \rev{\textbf{Dataset}} & & \rev{R@1} & \rev{R@3} & \rev{R@5} & & \rev{R@1} & \rev{R@3} & \rev{R@5} & & \rev{R@1} & \rev{R@3} & \rev{R@5} & & \rev{R@1} & \rev{R@3} & \rev{R@5} \\
\midrule
\rev{CLIP RN50} & \rev{OpenAI WIT} & & \rev{33.05} & \rev{51.17} & \rev{59.21} & & \rev{44.73} & \rev{61.32} & \rev{69.05} & & \rev{32.53} & \rev{59.96} & \rev{58.89} & & \rev{44.67} & \rev{61.43} & \rev{68.65} \\
\rev{CLIP ViT-B/32} & \rev{OpenAI WIT} & & \rev{34.70} & \rev{53.48} & \rev{61.31} & & \rev{46.73} & \rev{.63.26} & \rev{70.49} & & \rev{34.20} & \rev{52.73} & \rev{60.94} & & \rev{46.30} & \rev{62.62} & \rev{69.99} \\
\rev{OpenCLIP ViT-B/32} & \rev{LAION-400M} & & \rev{42.62} & \rev{62.31} & \rev{69.67} & & \rev{53.66} & \rev{69.71} & \rev{76.24} & & \rev{42.07} & \rev{61.74} & \rev{69.26} & & \rev{53.04} & \rev{69.06} & \rev{75.88} \\
\rev{OpenCLIP ViT-B/32} & \rev{LAION-2B} & & \rev{48.67} & \rev{67.68} & \rev{74.77} & & \rev{58.39} & \rev{73.76} & \rev{80.07} & & \rev{47.83} & \rev{67.25} & \rev{74.22} & & \rev{58.24} & \rev{73.60} & \rev{79.53} \\
\bottomrule
\end{tabular}
}
\vspace{-0.1cm}
\end{table}

\rev{Results are reported in Table~\ref{tab:retrieval}, using the exact pair and intra-cluster retrieval metrics and considering validation and test sets containing Stable Diffusion v1.4 and v2.0 images.} Surprisingly, retrieving the exact caption used to generate an image is not always easy, and the process is successful only in 40\% of the cases when selecting a proper backbone. Even when considering all captions of the same clusters as positives, moreover, we observe a recall@1 of around 50\%, again highlighting the difficulty of the task. \rev{The results are slightly higher when performing the experiment on the COCOFake version with Stable Diffusion v2.0 images, achieving 48.67\% and 58.39\% in terms of exact pair and intra-cluster retrieval on the validation set of the dataset. This suggests that the 2.0 version of Stable Diffusion can generate images more semantically aligned with the corresponding captions than the 1.4 version, probably due to the more powerful text encoder used in Stable Diffusion v2.0 (\ie, OpenCLIP ViT-H/14).} Nonetheless, these results point out that current generators produce images with partially altered semantic features, and are also in line with the previous observation that contrastive-based extractors percolate low-level features. 

\begin{table*}[t]
\caption{\rev{AUC and accuracy results on the semantic space $S$ and on the style space $T$. These results are obtained by training on the COCOFake training set with Stable Diffusion v1.4 images under the disentanglement setting and evaluating on test set of the COCOFake dataset, using data extracted from both Stable Diffusion v1.4 and v2.0.}}
\label{tab:centroids2}
\centering
\small
\setlength{\tabcolsep}{.4em}
\resizebox{0.92\linewidth}{!}{
\begin{tabular}{lc c >{\color{black}}ccc c >{\color{black}}ccc}
\toprule
& & & \multicolumn{7}{c}{\textbf{Test Set (SD v1.4 $\rightarrow$ SD v1.4)}} \\
\cmidrule{4-10}
& & & Overall & Overall & Full Cluster & & Overall & Min Dist. & Max Dist. \\
\textbf{Backbone} & \textbf{Dataset} & & AUC $S$ & Accuracy $S$ & Accuracy $S$ & & AUC $T$ & Accuracy $T$ & Accuracy $T$ \\
\midrule
RN50 & ImageNet & & 74.93 & 62.96 & 8.64 & & 98.45 & 0.42 & 89.08  \\ 
ViT-B/32 & ImageNet & & 68.19 & 64.04 & 8.46 & & 96.60 & 1.30 & 76.26  \\ 
\midrule
CLIP RN50 & OpenAI WIT & & 80.73 & 74.76 & 21.40 & & 99.87 & 0.00 & 98.46  \\ 
CLIP ViT-B/32 & OpenAI WIT & & 71.29 & 67.48 & 12.90 & & 99.74 & 0.20 & 98.14  \\
\midrule
OpenCLIP ViT-B/32 & LAION-400M & & 70.27 & 66.84 & 10.98 & & 99.45 & 0.10 & 94.48  \\
OpenCLIP ViT-B/32 & LAION-2B & & 78.00 & 72.62 & 17.32 & & 99.93 & 0.06 & 99.39  \\ 
\midrule
\midrule
& & & \multicolumn{7}{c}{\textbf{Test Set (SD v1.4 $\rightarrow$ SD v2.0)}} \\
\cmidrule{4-10}
& & & Overall & Overall & Full Cluster & & Overall & Min Dist. & Max Dist. \\
\textbf{Backbone} & \textbf{Dataset} & & AUC $S$ & Accuracy $S$ & Accuracy $S$ & & AUC $T$ & Accuracy $T$ & Accuracy $T$ \\
\midrule
RN50 & ImageNet & & 74.05 & 58.53 & 6.62 & & 98.15 & 0.52 & 89.48  \\ 
ViT-B/32 & ImageNet & & 68.46 & 63.00 & 8.86 & & 94.92 & 1.78 & 72.84  \\
\midrule
CLIP RN50 & OpenAI WIT & & 77.58 & 64.77 & 12.74 & & 99.71 & 0.12 & 96.42  \\
CLIP ViT-B/32 & OpenAI WIT & & 70.98 & 62.66 & 10.06 & & 99.30 & 0.26 & 94.60  \\ 
\midrule
OpenCLIP ViT-B/32 & LAION-400M & & 70.87 & 68.32 & 12.02 & & 98.25 & 0.52 & 83.98  \\
OpenCLIP ViT-B/32 & LAION-2B & & 76.49 & 71.92 & 16.98 & & 99.86 & 0.04 & 98.70  \\ 
\bottomrule
\end{tabular}
}
\vspace{-0.1cm}
\end{table*}

\subsection{Semantic-Style Disentangling Results} \label{sec:disentangle_exp}
We then turn our attention to evaluating the semantic-style disentanglement approach, in which we aim at training two separate embedding spaces, one storing semantic information and the second focusing on style information. \rev{We evaluate the semantic projection in terms of overall AUC and full cluster and overall classification accuracy, and the style projection in terms of overall AUC and minimum and maximum distance accuracy. Specifically, this is done by performing linear probing on top of the two disentangled projections $S$ and $T$, following the approach described in Sec.~\ref{sec:linear_probing}, and computing AUC and overall and full cluster accuracy scores. Instead, minimum and maximum distance accuracy are directly computed on the $T$ projection, to evaluate the relative spatial positions of the elements inside each cluster after disentangling semantics and style.} 

\rev{Results are reported in Table~\ref{tab:centroids2} and Table~\ref{tab:centroids3} on the COCOFake test set for all the aforementioned backbones, training the semantic-style disentanglement on the training set respectively with Stable Diffusion v1.4 and v2.0 images.} In both cases, we observe that, in the $T$ space which focuses on style, real and fake images can be properly distinguished, as the real image is always far apart from the generated ones. On the contrary, this does not happen in the $S$ space, which focuses on semantics, and in which all elements belonging to the same cluster are pulled together, independently of their authenticity. \rev{Still, the identification of deepfakes is feasible even in this more challenging space, although with lower AUC and accuracy scores (\ie, with an AUC up to 86\% and an accuracy of up to 80\%.)} As this corresponds to testing a more challenging generator that leaves fewer lower-level traces, we believe this result might offer interesting insights for future works. \rev{Similar but slightly lower results can also be observed when testing on images generated by the Stable Diffusion version not used during training, with an overall AUC up to 83\% and an overall accuracy up to 79\%. When instead considering the overall AUC computed over the $T$ projection, we can notice that the best results are above 99\% across almost all settings, thus confirming the proper distinction between real and fake images in the $T$ space.}

The structure of the two spaces can be further visualized in Fig.~\ref{fig:tsne}, in which we report 2D t-SNE visualizations~\citep{van2008visualizing} of the feature space of the OpenCLIP ViT-B/32 LAION-2B backbone, before and after disentanglement and for both Stable Diffusion v1.4 and Stable Diffusion v2.0. In the original embedding space, as provided by the backbone, real and generated samples appear to be mostly overlapped, even if we do not observe a complete overlap -- which is in line with the results presented in Table~\ref{tab:centroids} and Table~\ref{tab:linprob}. After the disentanglement, instead, the geometry of the $T$ and $S$ spaces appears completely different: the $T$ space clearly separates real and fake data (with the exception of a few outliers), while in the $S$ space we can observe a complete overlap between real and generated samples and a tendency to group into semantic clusters.

A closer visualization of the original feature space and the embedding spaces produced by the two projections is reported in Fig.~\ref{fig:supp_tsne}. In this case, we report, on each row, the relative positioning of eight sample clusters from the COCOFake test set with Stable Diffusion v1.4 images. As it can be seen, the two proposed projections are again effective both in separating real and fake images and in promoting the clustering of images sharing similar semantics regardless of their authenticity.

\begin{table*}[t]
\caption{\rev{AUC and accuracy results on the semantic space $S$ and on the style space $T$. These results are obtained by training on the COCOFake training set with Stable Diffusion v2.0 images under the disentanglement setting and evaluating on test set of the COCOFake dataset, using data extracted from both Stable Diffusion v1.4 and v2.0.}}
\label{tab:centroids3}
\centering
\small
\setlength{\tabcolsep}{.4em}
\resizebox{0.92\linewidth}{!}{
\begin{tabular}{>{\color{black}}l>{\color{black}}c c >{\color{black}}c>{\color{black}}c>{\color{black}}c c >{\color{black}}c>{\color{black}}c>{\color{black}}c}
\toprule
& & & \multicolumn{7}{c}{\rev{\textbf{Test Set (SD v2.0 $\rightarrow$ SD v2.0)}}} \\
\cmidrule{4-10}
& & & Overall & Overall & Full Cluster & & Overall & Min Dist. & Max Dist. \\
\textbf{Backbone} & \textbf{Dataset} & & AUC $S$ & Accuracy $S$ & Accuracy $S$ & & AUC $T$ & Accuracy $T$ & Accuracy $T$ \\
\midrule
RN50 & ImageNet & & 79.30 & 68.01 & 13.04 & & 98.43 & 0.54 & 89.58 \\ 
ViT-B/32 & ImageNet & & 69.20 & 66.31 & 11.40 & & 95.80 & 1.94 & 72.94 \\ 
\midrule
CLIP RN50 & OpenAI WIT & & 85.54 & 80.71 & 31.92 & & 99.79 & 0.04 & 97.92 \\ 
CLIP ViT-B/32 & OpenAI WIT & & 74.51 & 68.98 & 14.20 & & 99.76 & 0.08 & 97.60 \\
\midrule
OpenCLIP ViT-B/32 & LAION-400M & & 72.64 & 68.51 & 12.80 & & 99.02 & 0.38 & 90.52 \\
OpenCLIP ViT-B/32 & LAION-2B & & 82.69 & 76.60 & 23.94 & & 99.87 & 0.04 & 99.20 \\ 
\midrule
\midrule
& & & \multicolumn{7}{c}{\rev{\textbf{Test Set (SD v2.0 $\rightarrow$ SD v1.4)}}} \\
\cmidrule{4-10}
& & & Overall & Overall & Full Cluster & & Overall & Min Dist. & Max Dist. \\
\textbf{Backbone} & \textbf{Dataset} & & AUC $S$ & Accuracy $S$ & Accuracy $S$ & & AUC $T$ & Accuracy $T$ & Accuracy $T$ \\
\midrule
RN50 & ImageNet & & 76.87 & 67.91 & 12.62 & & 97.54 & 0.60 & 83.70 \\ 
ViT-B/32 & ImageNet & & 67.36 & 65.77 & 10.16 & & 94.45 & 2.34 & 69.00 \\ 
\midrule
CLIP RN50 & OpenAI WIT & & 83.00 & 78.67 & 27.10 & & 99.76 & 0.06 & 97.66 \\ 
CLIP ViT-B/32 & OpenAI WIT & & 72.48 & 68.79 & 45.36 & & 99.73 & 0.05 & 97.88 \\
\midrule
OpenCLIP ViT-B/32 & LAION-400M & & 69.85 & 65.39 & 9.60 & & 99.32 & 0.10 & 94.14 \\
OpenCLIP ViT-B/32 & LAION-2B & & 82.58 & 78.76 & 26.44 & & 99.86 & 0.08 & 99.34 \\ 
\bottomrule
\end{tabular}
}
\vspace{-0.1cm}
\end{table*}

\subsection{\rev{Robustness Analysis to Image Transformations}}
\rev{As shown in recent literature~\citep{fernandez2022watermarking,lu2023assessment,cocchi2023unveiling}, in addition to evaluating deepfake detection methods in a standard setting, it is also important to assess their robustness to image transformations, which may cause a severe performance degradation in some cases. To this aim, we replicate the experiment described in Sec.~\ref{sec:disentangle_exp} by testing on real and fake images that have undergone one of three considered image transformation techniques (\ie, Gaussian blur, JPEG compression, and resize). Specifically, we consider the disentangled spaces trained on non-transformed Stable Diffusion v2.0 images and evaluate on the corresponding test set where one image transformation is applied to all real and fake images, using a kernel size of 3 for Gaussian blurring, an image compression rate of 60 for JPEG compression, and an image edge size equal to 64 pixels for resizing.} 

\rev{Results are shown in Table~\ref{tab:transformations} in terms of the previously described AUC and accuracy evaluation metrics. Notably, while all image transformations cause a slight deterioration in performance, applying JPEG compression or scaling images to a lower resolution leads to the most drastic degradation of the final results, especially considering the results on the $T$ space with an overall AUC of 89\% to 97\% for JPEG compression and 87\% to 94\% for resizing. Conversely, Gaussian blur does not significantly impact deepfake detection performance with an overall AUC above 98\%.}

\begin{figure*}[t]
\centering
\footnotesize
\setlength{\tabcolsep}{.35em}
\resizebox{\linewidth}{!}{
\begin{tabular}{c ccc}
& \textbf{Original Features} & \textbf{Projected Features on $T$} & \textbf{Projected Features on $S$} \\
\addlinespace[0.08cm]
\rotatebox{90}{\parbox[t]{1.35in}{\hspace*{\fill}\textbf{Stable Diffusion v1.4}\hspace*{\fill}}} & 
\includegraphics[height=0.25\linewidth]{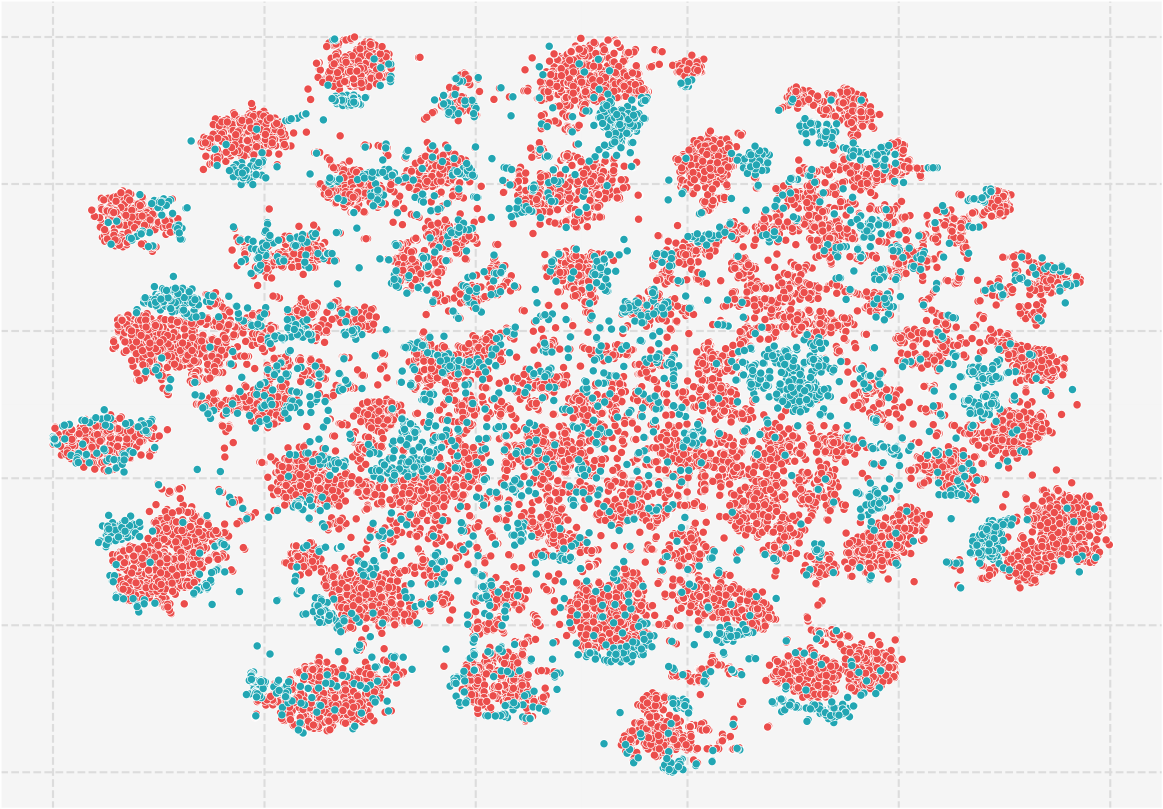} &
\includegraphics[height=0.25\linewidth]{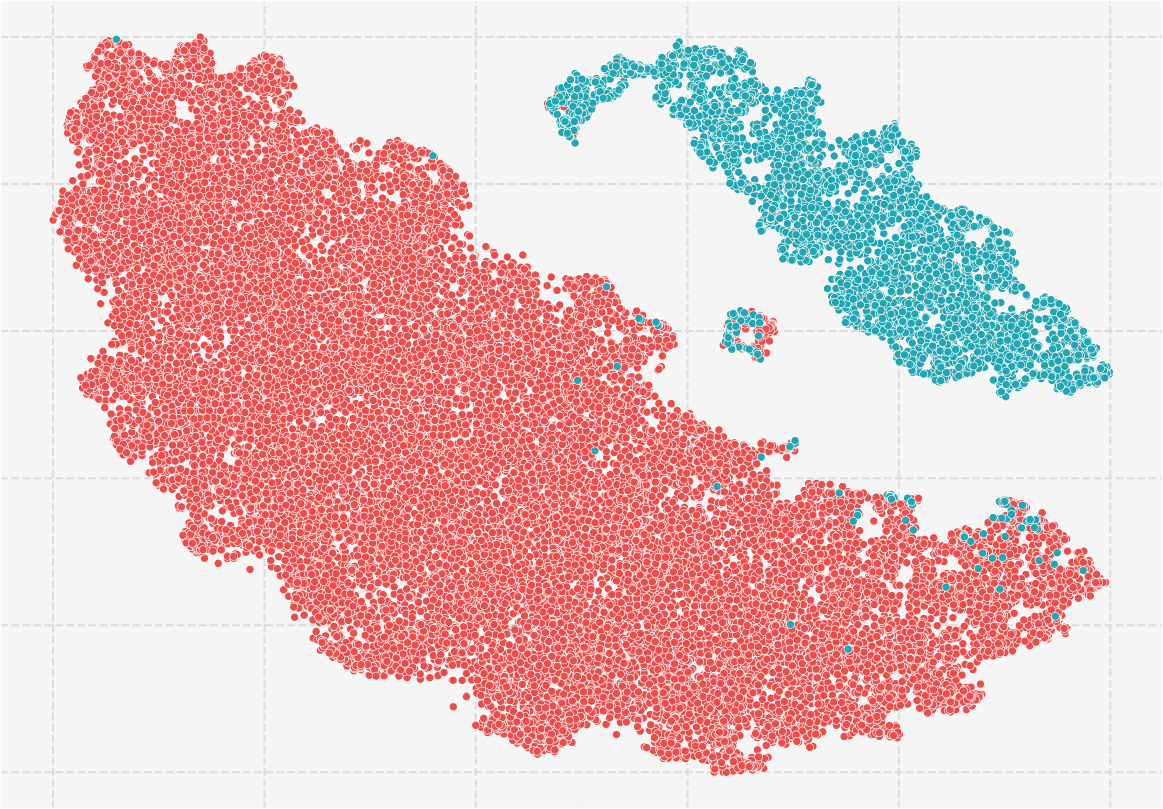} &
\includegraphics[height=0.25\linewidth]{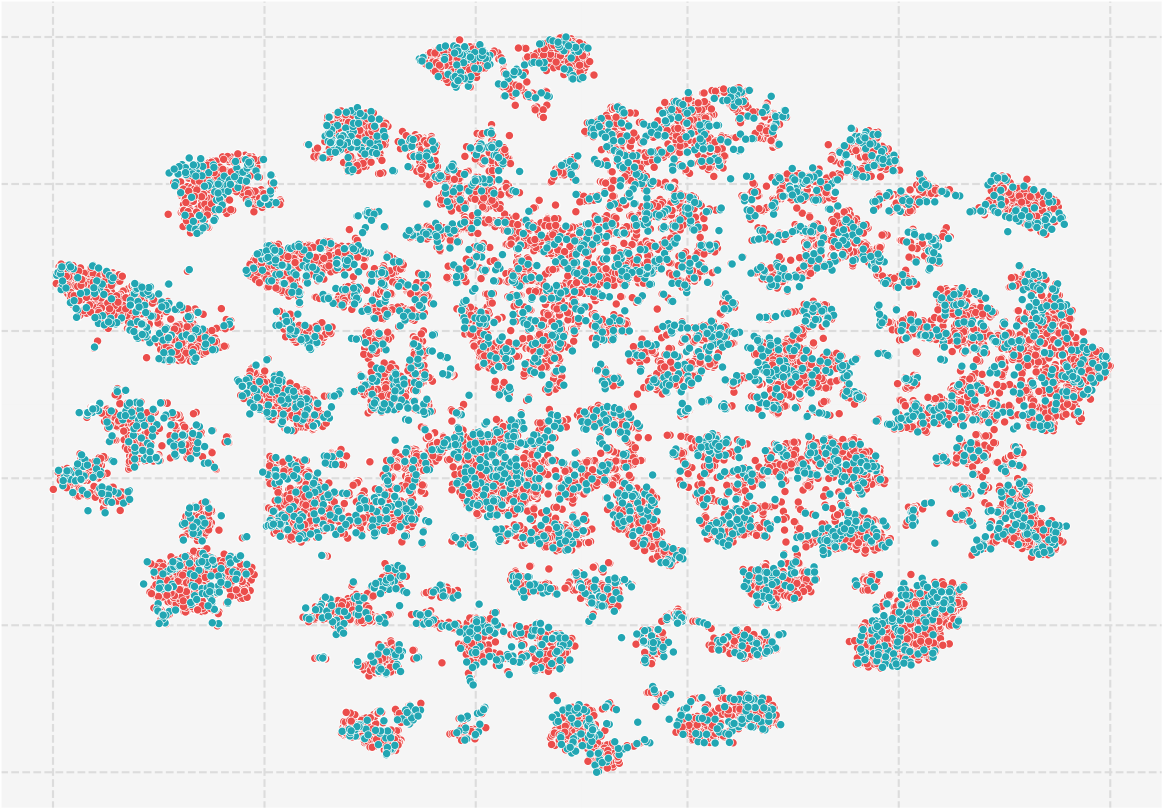} \\
\rotatebox{90}{\parbox[t]{1.35in}{\hspace*{\fill}\textbf{Stable Diffusion v2.0}\hspace*{\fill}}} & 
\includegraphics[height=0.25\linewidth]{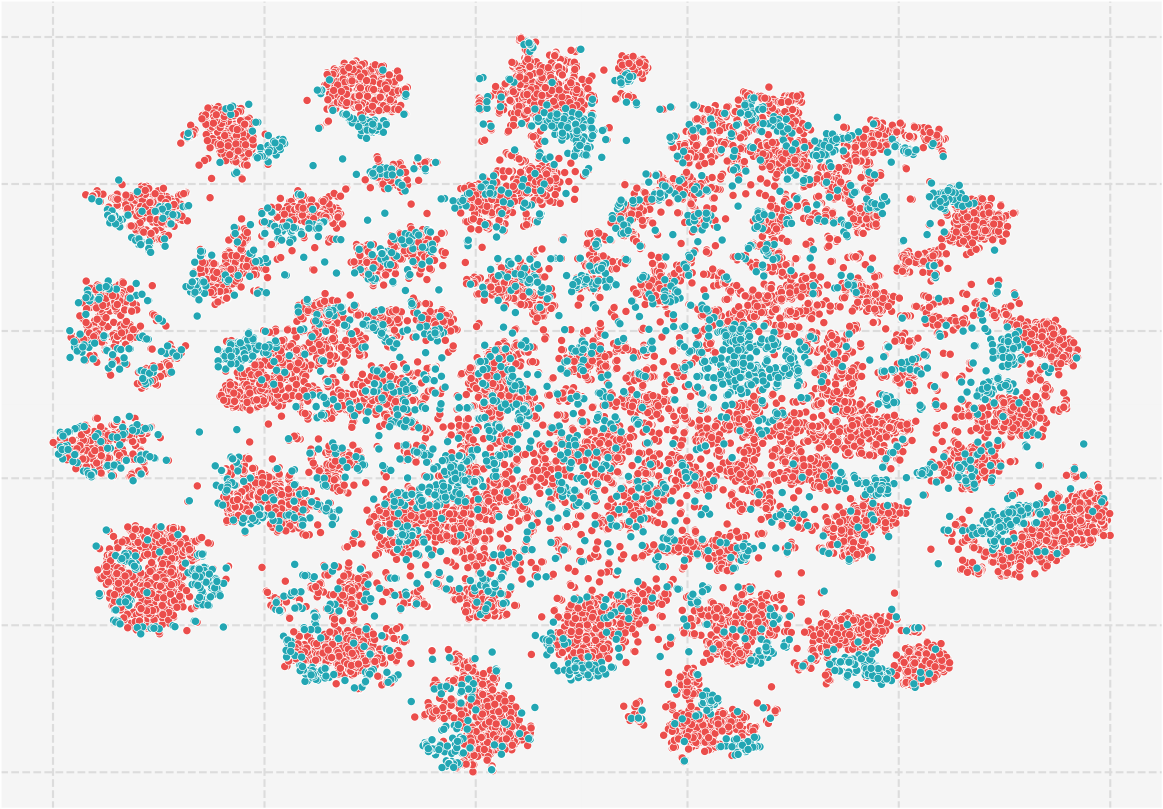} &
\includegraphics[height=0.25\linewidth]{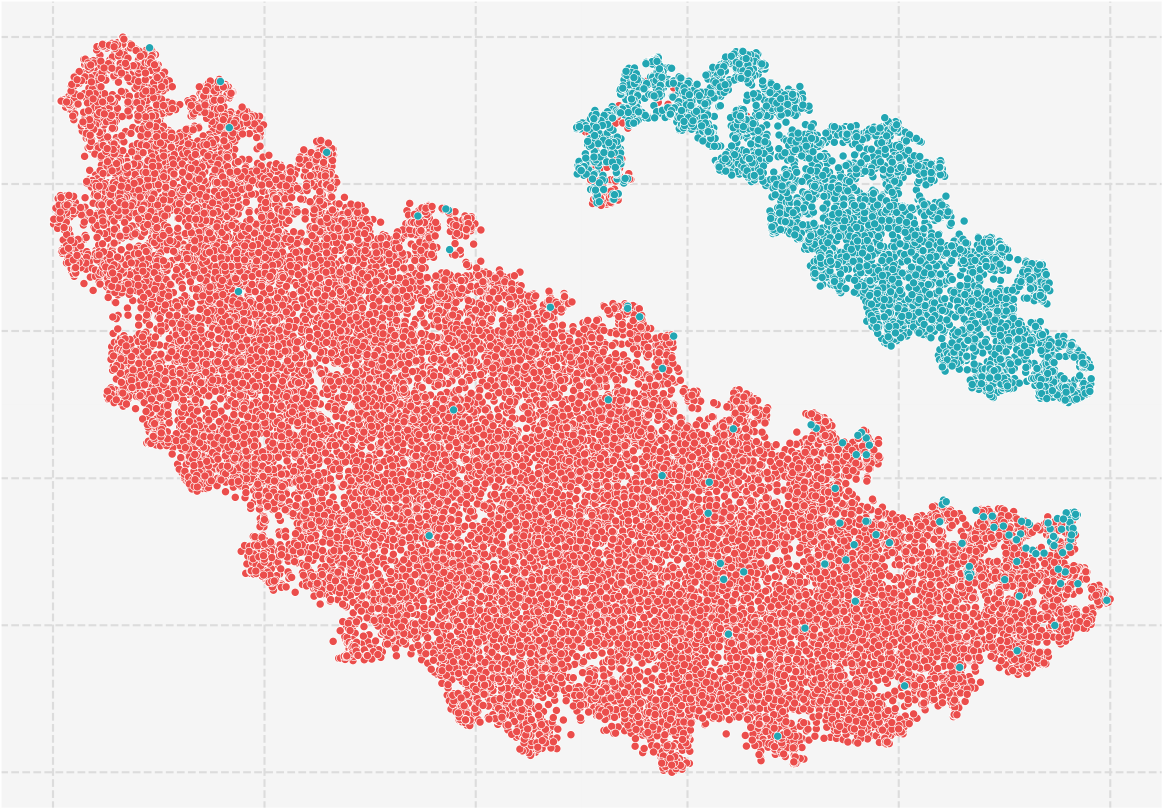} &
\includegraphics[height=0.25\linewidth]{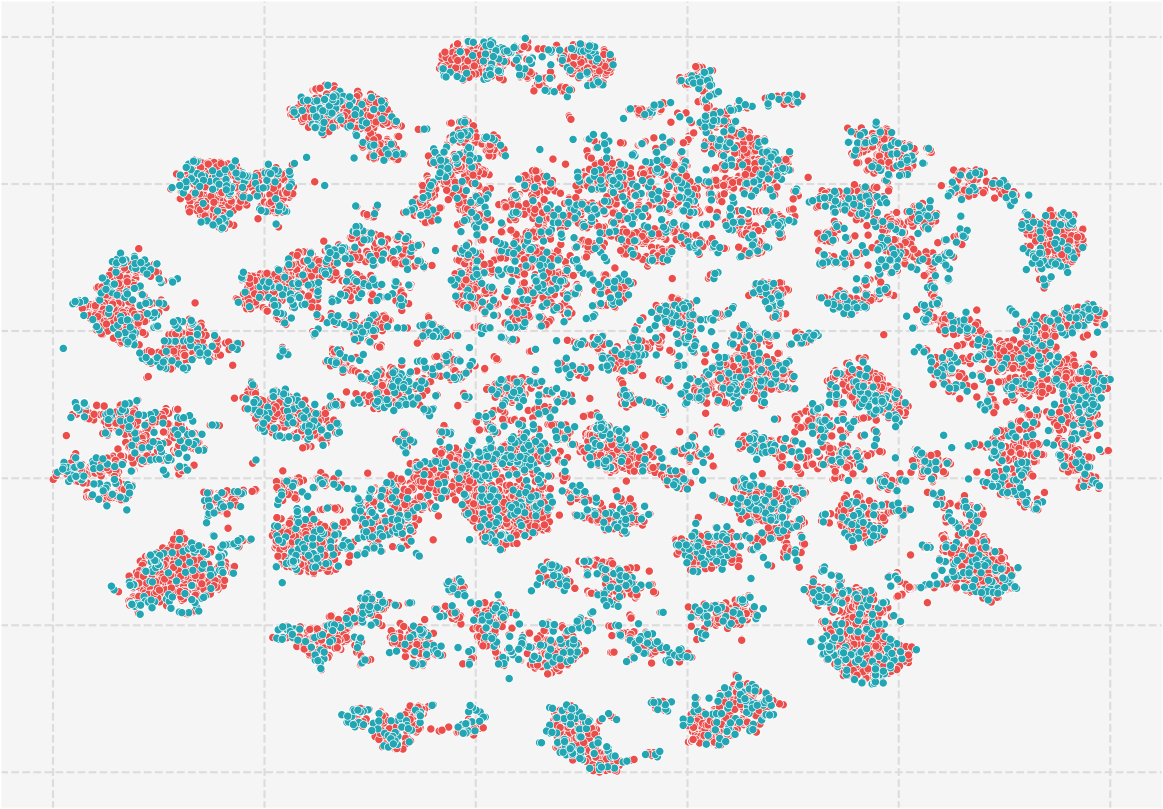} \\
\end{tabular}
}
\vspace{-0.1cm}
\caption{t-SNE visualizations over the validation set using the original visual features from the OpenCLIP ViT-B/32 LAION-2B backbone (left), the features projected on the $T$ space (style) after disentanglement (middle), and the features projected on the $S$ space (semantics) after disentanglement (right), using Stable Diffusion v1.4 (top) and v2.0 (bottom). Red dots indicate fake images, blue dots indicate real images.}
\label{fig:tsne}
\vspace{-0.1cm}
\end{figure*}

\subsection{\rev{Comparison with Other Methods}}
\rev{Finally, we compare our results with existing deepfake detection methods specifically tailored to recognize fake images from GAN-based generators. Specifically, we include in the comparison the models proposed by Wang~\etal~\citep{wang2020cnn} which are based on a ResNet-50 model trained with different image transformations (\ie, Gaussian blur and JPEG compression) and DetectGAN~\citep{mandelli2022detecting} based on an ensemble of different CNNs. For both competitors, we use the pre-trained weights downloaded from the official repositories provided by the authors.}

\rev{Table~\ref{tab:comparison} reports the results in terms of overall AUC, overall accuracy, and full cluster accuracy on the validation and test sets of COCOFake, using images generated by both versions of Stable Diffusion. Our results are obtained after disentangling semantics and style and by performing linear probing on the style space $T$ which is in charge of distinguishing real and fake images. As it can be seen, both competitors fail to effectively discriminate fake images from real ones with an overall AUC around 40\% for the model proposed in~\citep{wang2020cnn} and 55\% for the DetectGAN approach~\citep{mandelli2022detecting}, when tested on Stable Diffusion v1.4 images. On the contrary, all versions of our model achieve AUC scores greater than 99\% confirming the effectiveness of the $T$ space in correctly detecting deepfakes.}

\begin{figure}[!t]
\centering
\footnotesize
\setlength{\tabcolsep}{.5em}
\resizebox{\hsize}{!}{
\begin{tabular}{ccc}
\textbf{Original Features} & \textbf{Projected Features on $T$} & \textbf{Projected Features on $S$} \\
\addlinespace[0.08cm]
\includegraphics[height=0.25\linewidth]{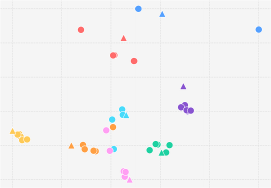} &
\includegraphics[height=0.25\linewidth]{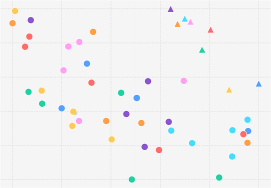} &
\includegraphics[height=0.25\linewidth]{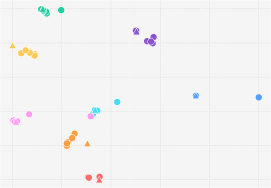} \\
\includegraphics[height=0.25\linewidth]{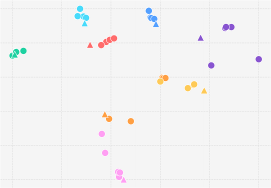} &
\includegraphics[height=0.25\linewidth]{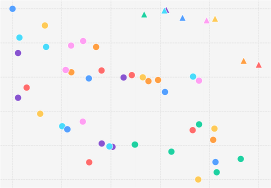} &
\includegraphics[height=0.25\linewidth]{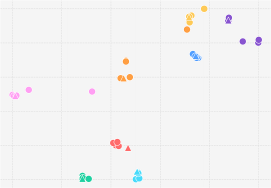} \\
\includegraphics[height=0.25\linewidth]{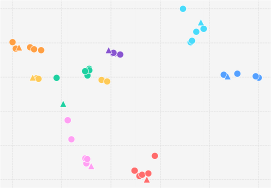} &
\includegraphics[height=0.25\linewidth]{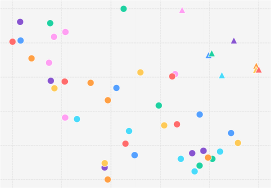} &
\includegraphics[height=0.25\linewidth]{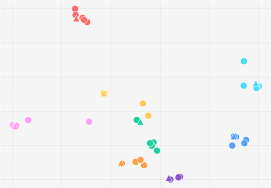} \\
\includegraphics[height=0.25\linewidth]{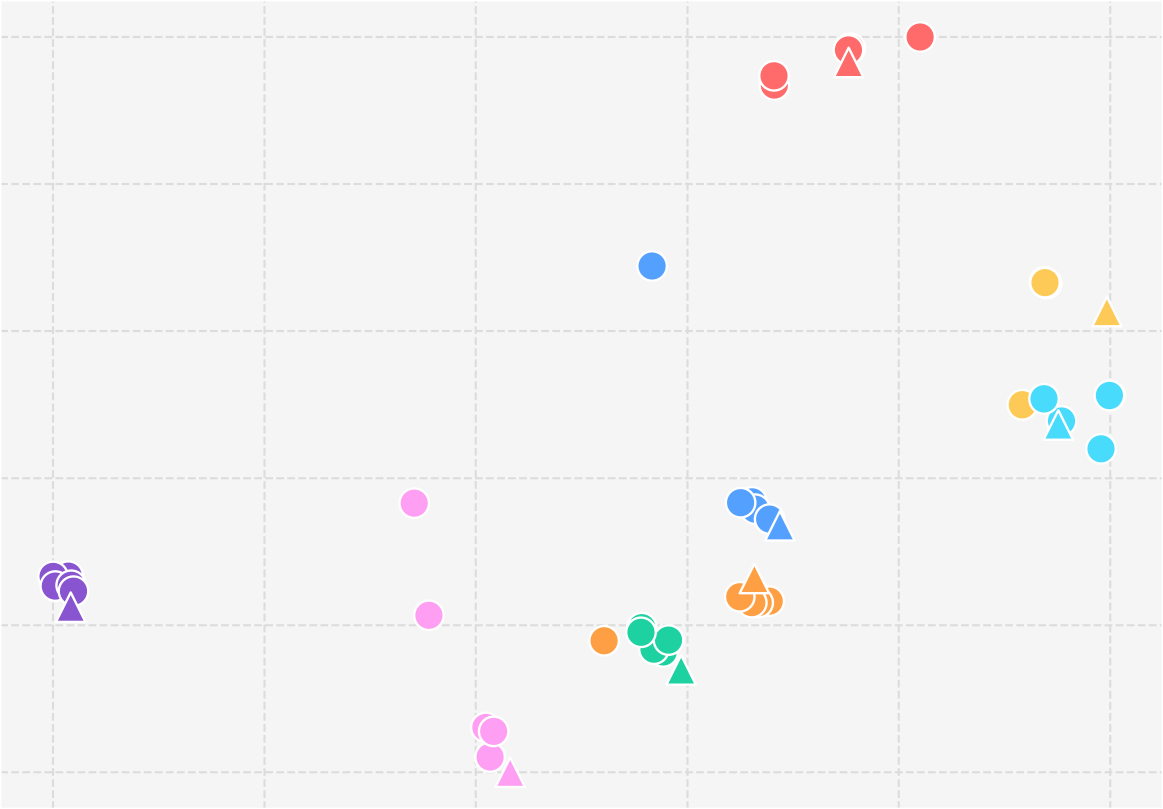} &
\includegraphics[height=0.25\linewidth]{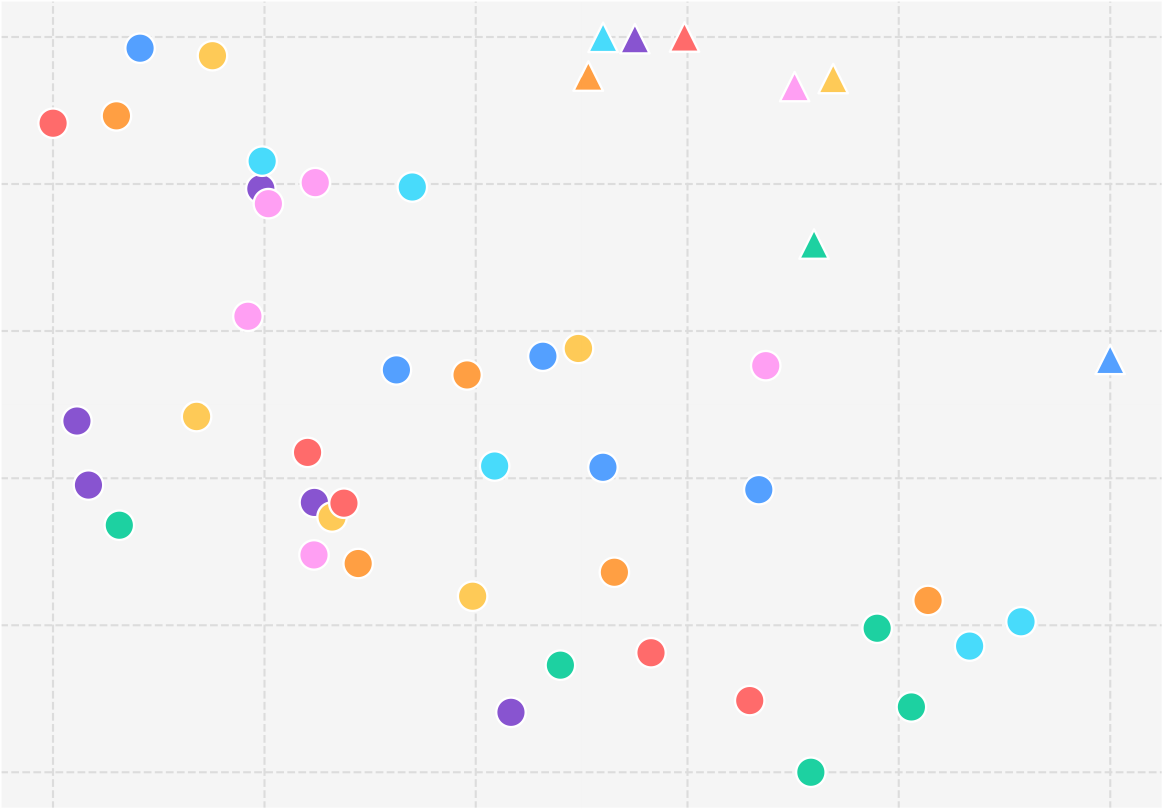} &
\includegraphics[height=0.25\linewidth]{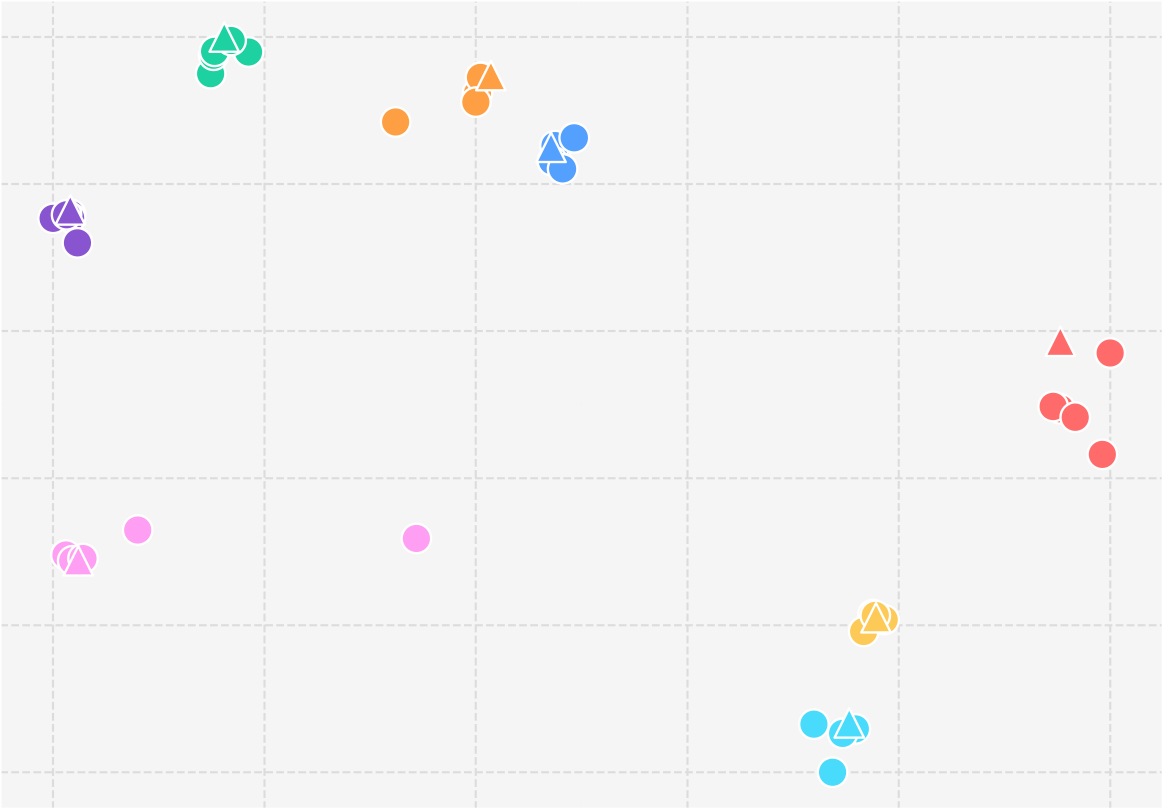} \\
\end{tabular}
}
\vspace{-0.1cm}
 \caption{t-SNE visualizations on sampled clusters from the Stable Diffusion v1.4 test set using features extracted from the OpenCLIP ViT-B/32 architecture pre-trained on LAION-2B. We report the original features from the visual backbone (left), the features projected on the $T$ space (style) after disentanglement (middle), and the features projected on the $S$ space (semantics) after disentanglement (right). Dots indicate fake images, triangles indicate real images. Images from the same cluster are shown with the same color.}
\label{fig:supp_tsne}
\vspace{-0.1cm}
\end{figure}

\begin{table*}[t]
\caption{\rev{AUC and accuracy results on the semantic space $S$ and on the style space $T$. These results are obtained by training on the COCOFake training set with Stable Diffusion v2.0 images under the disentanglement setting and evaluating on test set of the COCOFake dataset, using different image transformations.}}
\label{tab:transformations}
\centering
\small
\setlength{\tabcolsep}{.4em}
\resizebox{0.92\linewidth}{!}{
\begin{tabular}{>{\color{black}}l>{\color{black}}c c >{\color{black}}c>{\color{black}}c>{\color{black}}c c >{\color{black}}c>{\color{black}}c>{\color{black}}c}
\toprule
& & & \multicolumn{7}{c}{\rev{\textbf{Gaussian Blur (SD v2.0 $\rightarrow$ SD v2.0)}}} \\
\cmidrule{4-10}
& & & Overall & Overall & Full Cluster & & Overall & Min Dist. & Max Dist. \\
\textbf{Backbone} & \textbf{Dataset} & & AUC $S$ & Accuracy $S$ & Accuracy $S$ & & AUC $T$ & Accuracy $T$ & Accuracy $T$ \\
\midrule
CLIP RN50 & OpenAI WIT & & 77.44 & 74.28 & 19.42 & & 99.26 & 0.12 & 90.96 \\ 
CLIP ViT-B/32 & OpenAI WIT & & 70.41 & 59.65 & 7.72 & & 99.48 & 0.16 & 94.70 \\
OpenCLIP ViT-B/32 & LAION-400M & & 71.20 & 68.75 & 12.52 & & 98.27 & 0.56 & 86.28 \\
OpenCLIP ViT-B/32 & LAION-2B & & 79.31 & 75.16 & 21.38 & & 99.80 & 0.12 & 98.50 \\ 
\midrule
& & & \multicolumn{7}{c}{\rev{\textbf{JPEG Compression (SD v2.0 $\rightarrow$ SD v2.0)}}} \\
\cmidrule{4-10}
& & & Overall & Overall & Full Cluster & & Overall & Min Dist. & Max Dist. \\
\textbf{Backbone} & \textbf{Dataset} & & AUC $S$ & Accuracy $S$ & Accuracy $S$ & & AUC $T$ & Accuracy $T$ & Accuracy $T$ \\
\midrule
CLIP RN50 & OpenAI WIT & & 61.64 & 55.05 & 5.10 & & 88.60 & 3.62 & 57.62 \\ 
CLIP ViT-B/32 & OpenAI WIT & & 64.77 & 57.14 & 6.00 & & 89.38 & 4.04 & 54.30 \\
OpenCLIP ViT-B/32 & LAION-400M & & 69.22 & 62.01 & 8.26 & & 96.97 & 0.82 & 82.56 \\
OpenCLIP ViT-B/32 & LAION-2B & & 69.06 & 69.75 & 13.62 & & 93.32 & 2.28 & 65.66 \\ 
\midrule
& & & \multicolumn{7}{c}{\rev{\textbf{Resize (SD v2.0 $\rightarrow$ SD v2.0)}}} \\
\cmidrule{4-10}
& & & Overall & Overall & Full Cluster & & Overall & Min Dist. & Max Dist. \\
\textbf{Backbone} & \textbf{Dataset} & & AUC $S$ & Accuracy $S$ & Accuracy $S$ & & AUC $T$ & Accuracy $T$ & Accuracy $T$ \\
\midrule
CLIP RN50 & OpenAI WIT & & 62.75 & 70.05 & 10.50 & & 87.70 & 3.20 & 56.82 \\ 
CLIP ViT-B/32 & OpenAI WIT & & 67.78 & 31.70 & 0.36 & & 90.85 & 2.70 & 62.28 \\
OpenCLIP ViT-B/32 & LAION-400M & & 71.61 & 41.34 & 1.74 & & 94.32 & 2.16 & 72.50 \\
OpenCLIP ViT-B/32 & LAION-2B & & 75.12 & 30.71 & 0.70 & & 86.72 & 6.00 & 40.54 \\ 
\bottomrule
\end{tabular}
}
\vspace{-0.1cm}
\end{table*}

\begin{table*}[t]
\caption{\rev{Comparison with existing deepfake detection methods in terms of overall AUC, overall accuracy, and full cluster accuracy. Our results are obtained by performing linear probing on the style space $T$. Results are reported on the validation and test sets of COCOFake, using images extracted from both Stable Diffusion v1.4 and v2.0.}}
\label{tab:comparison}
\centering
\small
\setlength{\tabcolsep}{.4em}
\resizebox{0.97\linewidth}{!}{
\begin{tabular}{>{\color{black}}lc >{\color{black}}c>{\color{black}}c>{\color{black}}c c >{\color{black}}c>{\color{black}}c>{\color{black}}c}
\toprule
& & \multicolumn{3}{c}{\rev{\textbf{Validation Set (SD v1.4)}}} & & \multicolumn{3}{c}{\rev{\textbf{Test Set (SD v1.4)}}}  \\
\cmidrule{3-5} \cmidrule{7-9}
& & Overall & Overall & Full Cluster & & Overall & Overall & Full Cluster \\
\textbf{Model} & & AUC & Accuracy & Accuracy & & AUC & Accuracy & Accuracy \\
\midrule
Wang~\etal~(RN50 Blur+JPEG 0.5)~\citep{wang2020cnn} & & 40.61 & 83.26 & 0.34 & & 41.29 & 83.26 & 0.48 \\ 
Mandelli~\etal~(DetectGAN)~\citep{mandelli2022detecting} & & 54.55 & 83.12 & 4.78 & & 54.84 & 83.09 & 5.06 \\ 
\midrule
\textbf{Ours (CLIP RN50)} & & 99.85 & 98.79 & 93.32 & & 99.87 & 98.89 & 93.90 \\
\textbf{Ours (CLIP ViT-B/32)} & & 99.79 & 98.63 & 92.26 & & 99.74 & 98.47 & 91.58 \\
\textbf{Ours (OpenCLIP ViT-B/32 - LAION-400M)} & & 99.44 & 97.08 & 84.34 & & 99.45 & 97.21 & 85.00 \\
\textbf{Ours (OpenCLIP ViT-B/32 - LAION-2B)} & & \textbf{99.93} & \textbf{99.44} & \textbf{96.68} & & \textbf{99.93} & \textbf{99.39} & \textbf{96.44} \\
\midrule
\midrule
& & \multicolumn{3}{c}{\rev{\textbf{Validation Set (SD v2.0)}}} & & \multicolumn{3}{c}{\rev{\textbf{Test Set (SD v2.0)}}}  \\
\cmidrule{3-5} \cmidrule{7-9}
& & Overall & Overall & Full Cluster & & Overall & Overall & Full Cluster \\
\textbf{Model} & & AUC & Accuracy & Accuracy & & AUC & Accuracy & Accuracy \\
\midrule
Wang~\etal~(RN50 Blur+JPEG 0.5)~\citep{wang2020cnn} & & 53.05 & 83.32 & 0.40 & & 53.53 & 83.35 & 0.48 \\ 
Mandelli~\etal~(DetectGAN)~\citep{mandelli2022detecting} & & 64.26 & 83.55 & 4.98 & & 64.79 & 83.55 & 5.28 \\ 
\midrule
\textbf{Ours (CLIP RN50)} & & 99.79 & 98.38 & 91.06 & & 99.82 & 98.29 & 90.80 \\ 
\textbf{Ours (CLIP ViT-B/32)} & & 99.76 & 98.29 & 90.50 & & 99.69 & 98.14 & 90.10 \\
\textbf{Ours (OpenCLIP ViT-B/32 - LAION-400M)} & & 99.02 & 97.31 & 85.30 & & 99.14 & 97.28 & 84.98 \\
\textbf{Ours (OpenCLIP ViT-B/32 - LAION-2B)} & & \textbf{99.87} & \textbf{99.26} & \textbf{95.68} & & \textbf{99.87} & \textbf{99.30} & \textbf{96.02} \\
\bottomrule
\end{tabular}
}
\vspace{-0.1cm}
\end{table*}

\section{Conclusion}
\label{sec:conclusion}
This paper proposes a multimodal setting for deepfake detection and analysis, in which real and generated images sharing the same semantics are paired into semantic clusters. In our setting, different semantic projections of a given image, expressed through captions, are employed to generate fake images. Employing the popular Stable Diffusion model as generator, we investigated the performance of contrastive and classification-based visual features, highlighting that diffusion-based deepfakes share common low-level features that make them easily identifiable. Further, we proposed an approach to disentangle semantic and perceptual information, based on supervised contrastive learning. Under this setting, we investigated the classification of authenticity in a semantic space in which low-level cues left by the generator are removed, thus tackling a more challenging scenario. \rev{As a complementary contribution, we also collected and released the COCOFake dataset, containing about 1.2M images generated from COCO using both Stable Diffusion 1.4 and 2.0.} We believe that our work can shed further light on the development of deepfake detection strategies, also in consideration of the constant evolution of generator models.

\begin{acks}
We acknowledge the CINECA award under the ISCRA initiative, for the availability of high-performance computing resources and support. This work has been supported by the Horizon Europe project ``European Lighthouse on Safe and Secure AI (ELSA)'' (HORIZON-CL4-2021-HUMAN-01-03), co-funded by the European Union, and by the PNRR project ``Future Artificial Intelligence Research (FAIR)'', co-funded by the Italian Ministry of University and Research.
\end{acks}

\bibliographystyle{ACM-Reference-Format}
\bibliography{bibliography}


\begin{thebibliography}{75}


\ifx \showCODEN    \undefined \def \showCODEN     #1{\unskip}     \fi
\ifx \showDOI      \undefined \def \showDOI       #1{#1}\fi
\ifx \showISBNx    \undefined \def \showISBNx     #1{\unskip}     \fi
\ifx \showISBNxiii \undefined \def \showISBNxiii  #1{\unskip}     \fi
\ifx \showISSN     \undefined \def \showISSN      #1{\unskip}     \fi
\ifx \showLCCN     \undefined \def \showLCCN      #1{\unskip}     \fi
\ifx \shownote     \undefined \def \shownote      #1{#1}          \fi
\ifx \showarticletitle \undefined \def \showarticletitle #1{#1}   \fi
\ifx \showURL      \undefined \def \showURL       {\relax}        \fi
\providecommand\bibfield[2]{#2}
\providecommand\bibinfo[2]{#2}
\providecommand\natexlab[1]{#1}
\providecommand\showeprint[2][]{arXiv:#2}

\bibitem[Barraco et~al\mbox{.}(2023)]%
        {barraco2023little}
\bibfield{author}{\bibinfo{person}{Manuele Barraco}, \bibinfo{person}{Sara
  Sarto}, \bibinfo{person}{Marcella Cornia}, \bibinfo{person}{Lorenzo Baraldi},
  {and} \bibinfo{person}{Rita Cucchiara}.} \bibinfo{year}{2023}\natexlab{}.
\newblock \showarticletitle{{With a Little Help from your own Past:
  Prototypical Memory Networks for Image Captioning}}. In
  \bibinfo{booktitle}{\emph{Proceedings of the IEEE/CVF International
  Conference on Computer Vision}}.
\newblock


\bibitem[Brundage et~al\mbox{.}(2018)]%
        {brundage2018malicious}
\bibfield{author}{\bibinfo{person}{Miles Brundage}, \bibinfo{person}{Shahar
  Avin}, \bibinfo{person}{Jack Clark}, \bibinfo{person}{Helen Toner},
  \bibinfo{person}{Peter Eckersley}, \bibinfo{person}{Ben Garfinkel},
  \bibinfo{person}{Allan Dafoe}, \bibinfo{person}{Paul Scharre},
  \bibinfo{person}{Thomas Zeitzoff}, \bibinfo{person}{Bobby Filar},
  {et~al\mbox{.}}} \bibinfo{year}{2018}\natexlab{}.
\newblock \showarticletitle{{The Malicious Use of Artificial Intelligence:
  Forecasting, Prevention, and Mitigation}}.
\newblock \bibinfo{journal}{\emph{arXiv preprint arXiv:1802.07228}}
  (\bibinfo{year}{2018}).
\newblock


\bibitem[Byrd et~al\mbox{.}(1995)]%
        {byrd1995limited}
\bibfield{author}{\bibinfo{person}{Richard~H Byrd}, \bibinfo{person}{Peihuang
  Lu}, \bibinfo{person}{Jorge Nocedal}, {and} \bibinfo{person}{Ciyou Zhu}.}
  \bibinfo{year}{1995}\natexlab{}.
\newblock \showarticletitle{A limited memory algorithm for bound constrained
  optimization}.
\newblock \bibinfo{journal}{\emph{SIAM Journal on Scientific Computing}}
  \bibinfo{volume}{16}, \bibinfo{number}{5} (\bibinfo{year}{1995}),
  \bibinfo{pages}{1190--1208}.
\newblock


\bibitem[Caffagni et~al\mbox{.}(2023)]%
        {caffagni2023synthcap}
\bibfield{author}{\bibinfo{person}{Davide Caffagni}, \bibinfo{person}{Manuele
  Barraco}, \bibinfo{person}{Marcella Cornia}, \bibinfo{person}{Lorenzo
  Baraldi}, {and} \bibinfo{person}{Rita Cucchiara}.}
  \bibinfo{year}{2023}\natexlab{}.
\newblock \showarticletitle{{SynthCap: Augmenting Transformers with Synthetic
  Data for Image Captioning}}. In \bibinfo{booktitle}{\emph{Proceedings of the
  International Conference on Image Analysis and Processing}}.
\newblock


\bibitem[Chai et~al\mbox{.}(2020)]%
        {chai2020makes}
\bibfield{author}{\bibinfo{person}{Lucy Chai}, \bibinfo{person}{David Bau},
  \bibinfo{person}{Ser-Nam Lim}, {and} \bibinfo{person}{Phillip Isola}.}
  \bibinfo{year}{2020}\natexlab{}.
\newblock \showarticletitle{{What Makes Fake Images Detectable? Understanding
  Properties that Generalize}}. In \bibinfo{booktitle}{\emph{Proceedings of the
  European Conference on Computer Vision}}.
\newblock


\bibitem[Chandrasegaran et~al\mbox{.}(2021)]%
        {chandrasegaran2021closer}
\bibfield{author}{\bibinfo{person}{Keshigeyan Chandrasegaran},
  \bibinfo{person}{Ngoc-Trung Tran}, {and} \bibinfo{person}{Ngai-Man Cheung}.}
  \bibinfo{year}{2021}\natexlab{}.
\newblock \showarticletitle{{A Closer Look at Fourier Spectrum Discrepancies
  for CNN-Generated Images Detection}}. In
  \bibinfo{booktitle}{\emph{Proceedings of the IEEE/CVF Conference on Computer
  Vision and Pattern Recognition}}.
\newblock


\bibitem[Cheng et~al\mbox{.}(2023)]%
        {cheng2023voice}
\bibfield{author}{\bibinfo{person}{Harry Cheng}, \bibinfo{person}{Yangyang
  Guo}, \bibinfo{person}{Tianyi Wang}, \bibinfo{person}{Qi Li},
  \bibinfo{person}{Xiaojun Chang}, {and} \bibinfo{person}{Liqiang Nie}.}
  \bibinfo{year}{2023}\natexlab{}.
\newblock \showarticletitle{{Voice-Face Homogeneity Tells Deepfake}}.
\newblock \bibinfo{journal}{\emph{ACM Transactions on Multimedia Computing,
  Communications, and Applications}} \bibinfo{volume}{20}, \bibinfo{number}{3}
  (\bibinfo{year}{2023}), \bibinfo{pages}{1--22}.
\newblock


\bibitem[Chesney and Citron(2019)]%
        {chesney2019deepfakes}
\bibfield{author}{\bibinfo{person}{Robert Chesney} {and}
  \bibinfo{person}{Danielle Citron}.} \bibinfo{year}{2019}\natexlab{}.
\newblock \showarticletitle{Deepfakes and the new disinformation war: The
  coming age of post-truth geopolitics}.
\newblock \bibinfo{journal}{\emph{Foreign Affairs}}  \bibinfo{volume}{98}
  (\bibinfo{year}{2019}), \bibinfo{pages}{147}.
\newblock


\bibitem[Cocchi et~al\mbox{.}(2023)]%
        {cocchi2023unveiling}
\bibfield{author}{\bibinfo{person}{Federico Cocchi}, \bibinfo{person}{Lorenzo
  Baraldi}, \bibinfo{person}{Samuele Poppi}, \bibinfo{person}{Marcella Cornia},
  \bibinfo{person}{Lorenzo Baraldi}, {and} \bibinfo{person}{Rita Cucchiara}.}
  \bibinfo{year}{2023}\natexlab{}.
\newblock \showarticletitle{{Unveiling the Impact of Image Transformations on
  Deepfake Detection: An Experimental Analysis}}. In
  \bibinfo{booktitle}{\emph{Proceedings of the International Conference on
  Image Analysis and Processing}}.
\newblock


\bibitem[Corvi et~al\mbox{.}(2023)]%
        {corvi2022detection}
\bibfield{author}{\bibinfo{person}{Riccardo Corvi}, \bibinfo{person}{Davide
  Cozzolino}, \bibinfo{person}{Giada Zingarini}, \bibinfo{person}{Giovanni
  Poggi}, \bibinfo{person}{Koki Nagano}, {and} \bibinfo{person}{Luisa
  Verdoliva}.} \bibinfo{year}{2023}\natexlab{}.
\newblock \showarticletitle{{On The Detection of Synthetic Images Generated by
  Diffusion Models}}. In \bibinfo{booktitle}{\emph{Proceedings of the
  International Conference on Acoustics, Speech, and Signal Processing}}.
\newblock


\bibitem[Cozzolino et~al\mbox{.}(2021a)]%
        {cozzolino2021towards}
\bibfield{author}{\bibinfo{person}{Davide Cozzolino}, \bibinfo{person}{Diego
  Gragnaniello}, \bibinfo{person}{Giovanni Poggi}, {and} \bibinfo{person}{Luisa
  Verdoliva}.} \bibinfo{year}{2021}\natexlab{a}.
\newblock \showarticletitle{{Towards Universal GAN Image Detection}}. In
  \bibinfo{booktitle}{\emph{Proceedings of the International Conference on
  Visual Communications and Image Processing}}.
\newblock


\bibitem[Cozzolino et~al\mbox{.}(2021b)]%
        {cozzolino2021id}
\bibfield{author}{\bibinfo{person}{Davide Cozzolino}, \bibinfo{person}{Andreas
  R{\"o}ssler}, \bibinfo{person}{Justus Thies}, \bibinfo{person}{Matthias
  Nie{\ss}ner}, {and} \bibinfo{person}{Luisa Verdoliva}.}
  \bibinfo{year}{2021}\natexlab{b}.
\newblock \showarticletitle{{ID-Reveal: Identity-Aware DeepFake Video
  Detection}}. In \bibinfo{booktitle}{\emph{Proceedings of the IEEE/CVF
  International Conference on Computer Vision}}.
\newblock


\bibitem[Cozzolino et~al\mbox{.}(2018)]%
        {cozzolino2018forensictransfer}
\bibfield{author}{\bibinfo{person}{Davide Cozzolino}, \bibinfo{person}{Justus
  Thies}, \bibinfo{person}{Andreas R{\"o}ssler}, \bibinfo{person}{Christian
  Riess}, \bibinfo{person}{Matthias Nie{\ss}ner}, {and} \bibinfo{person}{Luisa
  Verdoliva}.} \bibinfo{year}{2018}\natexlab{}.
\newblock \showarticletitle{{ForensicTransfer: Weakly-supervised Domain
  Adaptation for Forgery Detection}}.
\newblock \bibinfo{journal}{\emph{arXiv preprint arXiv:1812.02510}}
  (\bibinfo{year}{2018}).
\newblock


\bibitem[Cucchiara et~al\mbox{.}(2024)]%
        {cucchiara2024video}
\bibfield{author}{\bibinfo{person}{Rita Cucchiara}, \bibinfo{person}{Lorenzo
  Baraldi}, \bibinfo{person}{Marcella Cornia}, {and} \bibinfo{person}{Sara
  Sarto}.} \bibinfo{year}{2024}\natexlab{}.
\newblock \showarticletitle{{Video Surveillance and Privacy: A Solvable
  Paradox?}}
\newblock \bibinfo{journal}{\emph{Computer}} \bibinfo{volume}{57},
  \bibinfo{number}{3} (\bibinfo{year}{2024}), \bibinfo{pages}{91--100}.
\newblock


\bibitem[Dhariwal and Nichol(2021)]%
        {dhariwal2021diffusion}
\bibfield{author}{\bibinfo{person}{Prafulla Dhariwal} {and}
  \bibinfo{person}{Alexander Nichol}.} \bibinfo{year}{2021}\natexlab{}.
\newblock \showarticletitle{{Diffusion Models Beat GANs on Image Synthesis}}.
  In \bibinfo{booktitle}{\emph{Advances in Neural Information Processing
  Systems}}.
\newblock


\bibitem[Ding et~al\mbox{.}(2021)]%
        {ding2021cogview}
\bibfield{author}{\bibinfo{person}{Ming Ding}, \bibinfo{person}{Zhuoyi Yang},
  \bibinfo{person}{Wenyi Hong}, \bibinfo{person}{Wendi Zheng},
  \bibinfo{person}{Chang Zhou}, \bibinfo{person}{Da Yin},
  \bibinfo{person}{Junyang Lin}, \bibinfo{person}{Xu Zou},
  \bibinfo{person}{Zhou Shao}, \bibinfo{person}{Hongxia Yang}, {and}
  \bibinfo{person}{Jie Tang}.} \bibinfo{year}{2021}\natexlab{}.
\newblock \showarticletitle{{CogView: Mastering Text-to-Image Generation via
  Transformers}}. In \bibinfo{booktitle}{\emph{Advances in Neural Information
  Processing Systems}}.
\newblock


\bibitem[Dolhansky et~al\mbox{.}(2020)]%
        {dolhansky2020deepfake}
\bibfield{author}{\bibinfo{person}{Brian Dolhansky}, \bibinfo{person}{Joanna
  Bitton}, \bibinfo{person}{Ben Pflaum}, \bibinfo{person}{Jikuo Lu},
  \bibinfo{person}{Russ Howes}, \bibinfo{person}{Menglin Wang}, {and}
  \bibinfo{person}{Cristian~Canton Ferrer}.} \bibinfo{year}{2020}\natexlab{}.
\newblock \showarticletitle{{The DeepFake Detection Challenge (DFDC) Dataset}}.
\newblock \bibinfo{journal}{\emph{arXiv preprint arXiv:2006.07397}}
  (\bibinfo{year}{2020}).
\newblock


\bibitem[Dosovitskiy et~al\mbox{.}(2020)]%
        {dosovitskiy2020image}
\bibfield{author}{\bibinfo{person}{Alexey Dosovitskiy}, \bibinfo{person}{Lucas
  Beyer}, \bibinfo{person}{Alexander Kolesnikov}, \bibinfo{person}{Dirk
  Weissenborn}, \bibinfo{person}{Xiaohua Zhai}, \bibinfo{person}{Thomas
  Unterthiner}, \bibinfo{person}{Mostafa Dehghani}, \bibinfo{person}{Matthias
  Minderer}, \bibinfo{person}{Georg Heigold}, \bibinfo{person}{Sylvain Gelly},
  {et~al\mbox{.}}} \bibinfo{year}{2020}\natexlab{}.
\newblock \showarticletitle{An Image is Worth 16x16 Words: Transformers for
  Image Recognition at Scale}. In \bibinfo{booktitle}{\emph{Proceedings of the
  International Conference on Learning Representations}}.
\newblock


\bibitem[Durall et~al\mbox{.}(2020)]%
        {durall2020watch}
\bibfield{author}{\bibinfo{person}{Ricard Durall}, \bibinfo{person}{Margret
  Keuper}, {and} \bibinfo{person}{Janis Keuper}.}
  \bibinfo{year}{2020}\natexlab{}.
\newblock \showarticletitle{{Watch Your Up-Convolution: CNN Based Generative
  Deep Neural Networks Are Failing to Reproduce Spectral Distributions}}. In
  \bibinfo{booktitle}{\emph{Proceedings of the IEEE/CVF Conference on Computer
  Vision and Pattern Recognition}}.
\newblock


\bibitem[Dzanic et~al\mbox{.}(2020)]%
        {dzanic2020fourier}
\bibfield{author}{\bibinfo{person}{Tarik Dzanic}, \bibinfo{person}{Karan Shah},
  {and} \bibinfo{person}{Freddie Witherden}.} \bibinfo{year}{2020}\natexlab{}.
\newblock \showarticletitle{{Fourier Spectrum Discrepancies in Deep Network
  Generated Images}}. In \bibinfo{booktitle}{\emph{Advances in Neural
  Information Processing Systems}}.
\newblock


\bibitem[Fernandez et~al\mbox{.}(2022)]%
        {fernandez2022watermarking}
\bibfield{author}{\bibinfo{person}{Pierre Fernandez},
  \bibinfo{person}{Alexandre Sablayrolles}, \bibinfo{person}{Teddy Furon},
  \bibinfo{person}{Herv{\'e} J{\'e}gou}, {and} \bibinfo{person}{Matthijs
  Douze}.} \bibinfo{year}{2022}\natexlab{}.
\newblock \showarticletitle{{Watermarking Images in Self-Supervised Latent
  Spaces}}. In \bibinfo{booktitle}{\emph{Proceedings of the International
  Conference on Acoustics, Speech, and Signal Processing}}.
\newblock


\bibitem[Frank et~al\mbox{.}(2020)]%
        {frank2020leveraging}
\bibfield{author}{\bibinfo{person}{Joel Frank}, \bibinfo{person}{Thorsten
  Eisenhofer}, \bibinfo{person}{Lea Sch{\"o}nherr}, \bibinfo{person}{Asja
  Fischer}, \bibinfo{person}{Dorothea Kolossa}, {and} \bibinfo{person}{Thorsten
  Holz}.} \bibinfo{year}{2020}\natexlab{}.
\newblock \showarticletitle{{Leveraging Frequency Analysis for Deep Fake Image
  Recognition}}. In \bibinfo{booktitle}{\emph{Proceedings of the International
  Conference on Machine Learning}}.
\newblock


\bibitem[Gafni et~al\mbox{.}(2022)]%
        {gafni2022make}
\bibfield{author}{\bibinfo{person}{Oran Gafni}, \bibinfo{person}{Adam Polyak},
  \bibinfo{person}{Oron Ashual}, \bibinfo{person}{Shelly Sheynin},
  \bibinfo{person}{Devi Parikh}, {and} \bibinfo{person}{Yaniv Taigman}.}
  \bibinfo{year}{2022}\natexlab{}.
\newblock \showarticletitle{{Make-A-Scene: Scene-Based Text-to-Image Generation
  with Human Priors}}. In \bibinfo{booktitle}{\emph{Proceedings of the European
  Conference on Computer Vision}}.
\newblock


\bibitem[Girish et~al\mbox{.}(2021)]%
        {girish2021towards}
\bibfield{author}{\bibinfo{person}{Sharath Girish}, \bibinfo{person}{Saksham
  Suri}, \bibinfo{person}{Sai~Saketh Rambhatla}, {and} \bibinfo{person}{Abhinav
  Shrivastava}.} \bibinfo{year}{2021}\natexlab{}.
\newblock \showarticletitle{{Towards Discovery and Attribution of Open-World
  GAN Generated Images}}. In \bibinfo{booktitle}{\emph{Proceedings of the
  IEEE/CVF International Conference on Computer Vision}}.
\newblock


\bibitem[Goodfellow et~al\mbox{.}(2014)]%
        {goodfellow2014generative}
\bibfield{author}{\bibinfo{person}{Ian~J Goodfellow}, \bibinfo{person}{Jean
  Pouget-Abadie}, \bibinfo{person}{Mehdi Mirza}, \bibinfo{person}{Bing Xu},
  \bibinfo{person}{David Warde-Farley}, \bibinfo{person}{Sherjil Ozair},
  \bibinfo{person}{Aaron~C Courville}, {and} \bibinfo{person}{Yoshua Bengio}.}
  \bibinfo{year}{2014}\natexlab{}.
\newblock \showarticletitle{{Generative Adversarial Nets}}. In
  \bibinfo{booktitle}{\emph{Advances in Neural Information Processing
  Systems}}.
\newblock


\bibitem[Gragnaniello et~al\mbox{.}(2021)]%
        {gragnaniello2021gan}
\bibfield{author}{\bibinfo{person}{Diego Gragnaniello}, \bibinfo{person}{Davide
  Cozzolino}, \bibinfo{person}{Francesco Marra}, \bibinfo{person}{Giovanni
  Poggi}, {and} \bibinfo{person}{Luisa Verdoliva}.}
  \bibinfo{year}{2021}\natexlab{}.
\newblock \showarticletitle{{Are GAN generated images easy to detect? A
  critical analysis of the state-of-the-art}}. In
  \bibinfo{booktitle}{\emph{Proceedings of the IEEE International Conference on
  Multimedia and Expo}}.
\newblock


\bibitem[Gu et~al\mbox{.}(2021)]%
        {gu2021spatiotemporal}
\bibfield{author}{\bibinfo{person}{Zhihao Gu}, \bibinfo{person}{Yang Chen},
  \bibinfo{person}{Taiping Yao}, \bibinfo{person}{Shouhong Ding},
  \bibinfo{person}{Jilin Li}, \bibinfo{person}{Feiyue Huang}, {and}
  \bibinfo{person}{Lizhuang Ma}.} \bibinfo{year}{2021}\natexlab{}.
\newblock \showarticletitle{{Spatiotemporal Inconsistency Learning for DeepFake
  Video Detection}}. In \bibinfo{booktitle}{\emph{Proceedings of the ACM
  International Conference on Multimedia}}.
\newblock


\bibitem[Han et~al\mbox{.}(2021)]%
        {han2021fighting}
\bibfield{author}{\bibinfo{person}{Bing Han}, \bibinfo{person}{Xiaoguang Han},
  \bibinfo{person}{Hua Zhang}, \bibinfo{person}{Jingzhi Li}, {and}
  \bibinfo{person}{Xiaochun Cao}.} \bibinfo{year}{2021}\natexlab{}.
\newblock \showarticletitle{{Fighting Fake News: Two Stream Network for
  Deepfake Detection via Learnable SRM}}.
\newblock \bibinfo{journal}{\emph{IEEE Transactions on Biometrics, Behavior,
  and Identity Science}} \bibinfo{volume}{3}, \bibinfo{number}{3}
  (\bibinfo{year}{2021}), \bibinfo{pages}{320--331}.
\newblock


\bibitem[Han et~al\mbox{.}(2023)]%
        {han2023sigma}
\bibfield{author}{\bibinfo{person}{Bing Han}, \bibinfo{person}{Jianshu Li},
  \bibinfo{person}{Wenqi Ren}, \bibinfo{person}{Man Luo}, \bibinfo{person}{Jian
  Liu}, {and} \bibinfo{person}{Xiaochun Cao}.} \bibinfo{year}{2023}\natexlab{}.
\newblock \showarticletitle{{SIGMA-DF: Single-Side Guided Meta-Learning for
  Deepfake Detection}}. In \bibinfo{booktitle}{\emph{Proceedings of the ACM
  International Conference on Multimedia Retrieval}}.
\newblock


\bibitem[Harris(2018)]%
        {harris2018deepfakes}
\bibfield{author}{\bibinfo{person}{Douglas Harris}.}
  \bibinfo{year}{2018}\natexlab{}.
\newblock \showarticletitle{{Deepfakes: False Pornography Is Here and the Law
  Cannot Protect You}}.
\newblock \bibinfo{journal}{\emph{Duke Law \& Technology Review}}
  \bibinfo{volume}{17} (\bibinfo{year}{2018}), \bibinfo{pages}{99}.
\newblock


\bibitem[He et~al\mbox{.}(2016)]%
        {he2016deep}
\bibfield{author}{\bibinfo{person}{Kaiming He}, \bibinfo{person}{Xiangyu
  Zhang}, \bibinfo{person}{Shaoqing Ren}, {and} \bibinfo{person}{Jian Sun}.}
  \bibinfo{year}{2016}\natexlab{}.
\newblock \showarticletitle{Deep residual learning for image recognition}. In
  \bibinfo{booktitle}{\emph{Proceedings of the IEEE/CVF Conference on Computer
  Vision and Pattern Recognition}}.
\newblock


\bibitem[Jiang et~al\mbox{.}(2020)]%
        {jiang2020deeperforensics}
\bibfield{author}{\bibinfo{person}{Liming Jiang}, \bibinfo{person}{Ren Li},
  \bibinfo{person}{Wayne Wu}, \bibinfo{person}{Chen Qian}, {and}
  \bibinfo{person}{Chen~Change Loy}.} \bibinfo{year}{2020}\natexlab{}.
\newblock \showarticletitle{{DeeperForensics-1.0: A Large-Scale Dataset for
  Real-World Face Forgery Detection}}. In \bibinfo{booktitle}{\emph{Proceedings
  of the IEEE/CVF Conference on Computer Vision and Pattern Recognition}}.
\newblock


\bibitem[Joslin and Hao(2020)]%
        {joslin2020attributing}
\bibfield{author}{\bibinfo{person}{Matthew Joslin} {and}
  \bibinfo{person}{Shuang Hao}.} \bibinfo{year}{2020}\natexlab{}.
\newblock \showarticletitle{{Attributing and Detecting Fake Images Generated by
  Known GANs}}. In \bibinfo{booktitle}{\emph{Proceedings of the IEEE Security
  and Privacy Workshops}}.
\newblock


\bibitem[Karpathy and Fei-Fei(2015)]%
        {karpathy2015deep}
\bibfield{author}{\bibinfo{person}{Andrej Karpathy} {and} \bibinfo{person}{Li
  Fei-Fei}.} \bibinfo{year}{2015}\natexlab{}.
\newblock \showarticletitle{Deep visual-semantic alignments for generating
  image descriptions}. In \bibinfo{booktitle}{\emph{Proceedings of the IEEE/CVF
  Conference on Computer Vision and Pattern Recognition}}.
\newblock


\bibitem[Karras et~al\mbox{.}(2019)]%
        {karras2019style}
\bibfield{author}{\bibinfo{person}{Tero Karras}, \bibinfo{person}{Samuli
  Laine}, {and} \bibinfo{person}{Timo Aila}.} \bibinfo{year}{2019}\natexlab{}.
\newblock \showarticletitle{{A Style-Based Generator Architecture for
  Generative Adversarial Networks}}. In \bibinfo{booktitle}{\emph{Proceedings
  of the IEEE/CVF Conference on Computer Vision and Pattern Recognition}}.
\newblock


\bibitem[Karras et~al\mbox{.}(2020)]%
        {karras2020analyzing}
\bibfield{author}{\bibinfo{person}{Tero Karras}, \bibinfo{person}{Samuli
  Laine}, \bibinfo{person}{Miika Aittala}, \bibinfo{person}{Janne Hellsten},
  \bibinfo{person}{Jaakko Lehtinen}, {and} \bibinfo{person}{Timo Aila}.}
  \bibinfo{year}{2020}\natexlab{}.
\newblock \showarticletitle{{Analyzing and Improving the Image Quality of
  StyleGAN}}. In \bibinfo{booktitle}{\emph{Proceedings of the IEEE/CVF
  Conference on Computer Vision and Pattern Recognition}}.
\newblock


\bibitem[Khayatkhoei and Elgammal(2022)]%
        {khayatkhoei2022spatial}
\bibfield{author}{\bibinfo{person}{Mahyar Khayatkhoei} {and}
  \bibinfo{person}{Ahmed Elgammal}.} \bibinfo{year}{2022}\natexlab{}.
\newblock \showarticletitle{{Spatial Frequency Bias in Convolutional Generative
  Adversarial Networks}}. In \bibinfo{booktitle}{\emph{Proceedings of the AAAI
  Conference on Artificial Intelligence}}.
\newblock


\bibitem[Khosla et~al\mbox{.}(2020)]%
        {khosla2020supervised}
\bibfield{author}{\bibinfo{person}{Prannay Khosla}, \bibinfo{person}{Piotr
  Teterwak}, \bibinfo{person}{Chen Wang}, \bibinfo{person}{Aaron Sarna},
  \bibinfo{person}{Yonglong Tian}, \bibinfo{person}{Phillip Isola},
  \bibinfo{person}{Aaron Maschinot}, \bibinfo{person}{Ce Liu}, {and}
  \bibinfo{person}{Dilip Krishnan}.} \bibinfo{year}{2020}\natexlab{}.
\newblock \showarticletitle{Supervised contrastive learning}. In
  \bibinfo{booktitle}{\emph{Advances in Neural Information Processing
  Systems}}.
\newblock


\bibitem[Kingma and Dhariwal(2018)]%
        {kingma2018glow}
\bibfield{author}{\bibinfo{person}{Durk~P Kingma} {and}
  \bibinfo{person}{Prafulla Dhariwal}.} \bibinfo{year}{2018}\natexlab{}.
\newblock \showarticletitle{{Glow: Generative Flow with Invertible 1x1
  Convolutions}}. In \bibinfo{booktitle}{\emph{Advances in Neural Information
  Processing Systems}}.
\newblock


\bibitem[Li and Lyu(2018)]%
        {li2018exposing}
\bibfield{author}{\bibinfo{person}{Yuezun Li} {and} \bibinfo{person}{Siwei
  Lyu}.} \bibinfo{year}{2018}\natexlab{}.
\newblock \showarticletitle{{Exposing DeepFake Videos By Detecting Face Warping
  Artifacts}}. In \bibinfo{booktitle}{\emph{Proceedings of the IEEE/CVF
  Conference on Computer Vision and Pattern Recognition Workshops}}.
\newblock


\bibitem[Li et~al\mbox{.}(2020)]%
        {li2020celeb}
\bibfield{author}{\bibinfo{person}{Yuezun Li}, \bibinfo{person}{Xin Yang},
  \bibinfo{person}{Pu Sun}, \bibinfo{person}{Honggang Qi}, {and}
  \bibinfo{person}{Siwei Lyu}.} \bibinfo{year}{2020}\natexlab{}.
\newblock \showarticletitle{[Celeb-DF: A Large-Scale Challenging Dataset for
  DeepFake Forensics]}. In \bibinfo{booktitle}{\emph{Proceedings of the
  IEEE/CVF Conference on Computer Vision and Pattern Recognition}}.
\newblock


\bibitem[Lin et~al\mbox{.}(2014)]%
        {lin2014microsoft}
\bibfield{author}{\bibinfo{person}{Tsung-Yi Lin}, \bibinfo{person}{Michael
  Maire}, \bibinfo{person}{Serge Belongie}, \bibinfo{person}{James Hays},
  \bibinfo{person}{Pietro Perona}, \bibinfo{person}{Deva Ramanan},
  \bibinfo{person}{Piotr Doll{\'a}r}, {and} \bibinfo{person}{C~Lawrence
  Zitnick}.} \bibinfo{year}{2014}\natexlab{}.
\newblock \showarticletitle{{Microsoft COCO: Common Objects in Context}}. In
  \bibinfo{booktitle}{\emph{Proceedings of the European Conference on Computer
  Vision}}.
\newblock


\bibitem[Loshchilov and Hutter(2018)]%
        {loshchilov2018decoupled}
\bibfield{author}{\bibinfo{person}{Ilya Loshchilov} {and}
  \bibinfo{person}{Frank Hutter}.} \bibinfo{year}{2018}\natexlab{}.
\newblock \showarticletitle{Decoupled Weight Decay Regularization}. In
  \bibinfo{booktitle}{\emph{Proceedings of the International Conference on
  Learning Representations}}.
\newblock


\bibitem[Lu and Ebrahimi(2024)]%
        {lu2023assessment}
\bibfield{author}{\bibinfo{person}{Yuhang Lu} {and} \bibinfo{person}{Touradj
  Ebrahimi}.} \bibinfo{year}{2024}\natexlab{}.
\newblock \showarticletitle{{Assessment Framework for Deepfake Detection in
  Real-World Situations}}.
\newblock \bibinfo{journal}{\emph{EURASIP Journal on Image and Video
  Processing}} \bibinfo{volume}{2024}, \bibinfo{number}{1}
  (\bibinfo{year}{2024}), \bibinfo{pages}{6}.
\newblock


\bibitem[Luo et~al\mbox{.}(2021)]%
        {luo2021generalizing}
\bibfield{author}{\bibinfo{person}{Yuchen Luo}, \bibinfo{person}{Yong Zhang},
  \bibinfo{person}{Junchi Yan}, {and} \bibinfo{person}{Wei Liu}.}
  \bibinfo{year}{2021}\natexlab{}.
\newblock \showarticletitle{{Generalizing Face Forgery Detection With
  High-Frequency Features}}. In \bibinfo{booktitle}{\emph{Proceedings of the
  IEEE/CVF Conference on Computer Vision and Pattern Recognition}}.
\newblock


\bibitem[Mandelli et~al\mbox{.}(2022)]%
        {mandelli2022detecting}
\bibfield{author}{\bibinfo{person}{Sara Mandelli}, \bibinfo{person}{Nicol{\`o}
  Bonettini}, \bibinfo{person}{Paolo Bestagini}, {and} \bibinfo{person}{Stefano
  Tubaro}.} \bibinfo{year}{2022}\natexlab{}.
\newblock \showarticletitle{{Detecting GAN-generated Images by Orthogonal
  Training of Multiple CNNs}}. In \bibinfo{booktitle}{\emph{Proceedings of the
  International Conference on Image Processing}}.
\newblock


\bibitem[Marra et~al\mbox{.}(2018)]%
        {marra2018detection}
\bibfield{author}{\bibinfo{person}{Francesco Marra}, \bibinfo{person}{Diego
  Gragnaniello}, \bibinfo{person}{Davide Cozzolino}, {and}
  \bibinfo{person}{Luisa Verdoliva}.} \bibinfo{year}{2018}\natexlab{}.
\newblock \showarticletitle{{Detection of GAN-Generated Fake Images over Social
  Networks}}. In \bibinfo{booktitle}{\emph{Proceedings of the IEEE Conference
  on Multimedia Information Processing and Retrieval}}.
\newblock


\bibitem[Mirza and Osindero(2014)]%
        {mirza2014conditional}
\bibfield{author}{\bibinfo{person}{Mehdi Mirza} {and} \bibinfo{person}{Simon
  Osindero}.} \bibinfo{year}{2014}\natexlab{}.
\newblock \showarticletitle{{Conditional Generative Adversarial Nets}}.
\newblock \bibinfo{journal}{\emph{arXiv preprint arXiv:1411.1784}}
  (\bibinfo{year}{2014}).
\newblock


\bibitem[Nichol et~al\mbox{.}(2021)]%
        {nichol2021glide}
\bibfield{author}{\bibinfo{person}{Alex Nichol}, \bibinfo{person}{Prafulla
  Dhariwal}, \bibinfo{person}{Aditya Ramesh}, \bibinfo{person}{Pranav Shyam},
  \bibinfo{person}{Pamela Mishkin}, \bibinfo{person}{Bob McGrew},
  \bibinfo{person}{Ilya Sutskever}, {and} \bibinfo{person}{Mark Chen}.}
  \bibinfo{year}{2021}\natexlab{}.
\newblock \showarticletitle{{GLIDE: Towards Photorealistic Image Generation and
  Editing with Text-Guided Diffusion Models}}.
\newblock \bibinfo{journal}{\emph{arXiv preprint arXiv:2112.10741}}
  (\bibinfo{year}{2021}).
\newblock


\bibitem[Poppi et~al\mbox{.}(2024)]%
        {poppi2023removing}
\bibfield{author}{\bibinfo{person}{Samuele Poppi}, \bibinfo{person}{Tobia
  Poppi}, \bibinfo{person}{Federico Cocchi}, \bibinfo{person}{Marcella Cornia},
  \bibinfo{person}{Lorenzo Baraldi}, {and} \bibinfo{person}{Rita Cucchiara}.}
  \bibinfo{year}{2024}\natexlab{}.
\newblock \showarticletitle{{Safe-CLIP: Removing NSFW Concepts from
  Vision-and-Language Models}}.
\newblock \bibinfo{journal}{\emph{arXiv preprint arXiv:2311.16254}}
  (\bibinfo{year}{2024}).
\newblock


\bibitem[Radford et~al\mbox{.}(2021)]%
        {radford2021learning}
\bibfield{author}{\bibinfo{person}{Alec Radford}, \bibinfo{person}{Jong~Wook
  Kim}, \bibinfo{person}{Chris Hallacy}, \bibinfo{person}{Aditya Ramesh},
  \bibinfo{person}{Gabriel Goh}, \bibinfo{person}{Sandhini Agarwal},
  \bibinfo{person}{Girish Sastry}, \bibinfo{person}{Amanda Askell},
  \bibinfo{person}{Pamela Mishkin}, \bibinfo{person}{Jack Clark},
  {et~al\mbox{.}}} \bibinfo{year}{2021}\natexlab{}.
\newblock \showarticletitle{Learning transferable visual models from natural
  language supervision}. In \bibinfo{booktitle}{\emph{Proceedings of the
  International Conference on Machine Learning}}.
\newblock


\bibitem[Ramesh et~al\mbox{.}(2022)]%
        {ramesh2022hierarchical}
\bibfield{author}{\bibinfo{person}{Aditya Ramesh}, \bibinfo{person}{Prafulla
  Dhariwal}, \bibinfo{person}{Alex Nichol}, \bibinfo{person}{Casey Chu}, {and}
  \bibinfo{person}{Mark Chen}.} \bibinfo{year}{2022}\natexlab{}.
\newblock \showarticletitle{{Hierarchical Text-Conditional Image Generation
  with CLIP Latents}}.
\newblock \bibinfo{journal}{\emph{arXiv preprint arXiv:2204.06125}}
  (\bibinfo{year}{2022}).
\newblock


\bibitem[Ramesh et~al\mbox{.}(2021)]%
        {ramesh2021zero}
\bibfield{author}{\bibinfo{person}{Aditya Ramesh}, \bibinfo{person}{Mikhail
  Pavlov}, \bibinfo{person}{Gabriel Goh}, \bibinfo{person}{Scott Gray},
  \bibinfo{person}{Chelsea Voss}, \bibinfo{person}{Alec Radford},
  \bibinfo{person}{Mark Chen}, {and} \bibinfo{person}{Ilya Sutskever}.}
  \bibinfo{year}{2021}\natexlab{}.
\newblock \showarticletitle{{Zero-Shot Text-to-Image Generation}}. In
  \bibinfo{booktitle}{\emph{Proceedings of the International Conference on
  Machine Learning}}.
\newblock


\bibitem[Ricker et~al\mbox{.}(2024)]%
        {ricker2022towards}
\bibfield{author}{\bibinfo{person}{Jonas Ricker}, \bibinfo{person}{Simon Damm},
  \bibinfo{person}{Thorsten Holz}, {and} \bibinfo{person}{Asja Fischer}.}
  \bibinfo{year}{2024}\natexlab{}.
\newblock \showarticletitle{{Towards the Detection of Diffusion Model
  Deepfakes}}. In \bibinfo{booktitle}{\emph{Proceedings of the International
  Joint Conference on Computer Vision, Imaging and Computer Graphics Theory and
  Applications}}.
\newblock


\bibitem[Rombach et~al\mbox{.}(2022)]%
        {rombach2022high}
\bibfield{author}{\bibinfo{person}{Robin Rombach}, \bibinfo{person}{Andreas
  Blattmann}, \bibinfo{person}{Dominik Lorenz}, \bibinfo{person}{Patrick
  Esser}, {and} \bibinfo{person}{Bj{\"o}rn Ommer}.}
  \bibinfo{year}{2022}\natexlab{}.
\newblock \showarticletitle{{High-Resolution Image Synthesis With Latent
  Diffusion Models}}. In \bibinfo{booktitle}{\emph{Proceedings of the IEEE/CVF
  Conference on Computer Vision and Pattern Recognition}}.
\newblock


\bibitem[Rossler et~al\mbox{.}(2019)]%
        {rossler2019faceforensics}
\bibfield{author}{\bibinfo{person}{Andreas Rossler}, \bibinfo{person}{Davide
  Cozzolino}, \bibinfo{person}{Luisa Verdoliva}, \bibinfo{person}{Christian
  Riess}, \bibinfo{person}{Justus Thies}, {and} \bibinfo{person}{Matthias
  Nie{\ss}ner}.} \bibinfo{year}{2019}\natexlab{}.
\newblock \showarticletitle{{FaceForensics++: Learning to Detect Manipulated
  Facial Images}}. In \bibinfo{booktitle}{\emph{Proceedings of the IEEE/CVF
  International Conference on Computer Vision}}.
\newblock


\bibitem[Russakovsky et~al\mbox{.}(2015)]%
        {russakovsky2015imagenet}
\bibfield{author}{\bibinfo{person}{Olga Russakovsky}, \bibinfo{person}{Jia
  Deng}, \bibinfo{person}{Hao Su}, \bibinfo{person}{Jonathan Krause},
  \bibinfo{person}{Sanjeev Satheesh}, \bibinfo{person}{Sean Ma},
  \bibinfo{person}{Zhiheng Huang}, \bibinfo{person}{Andrej Karpathy},
  \bibinfo{person}{Aditya Khosla}, \bibinfo{person}{Michael Bernstein},
  \bibinfo{person}{Alexander~C. Berg}, {and} \bibinfo{person}{Li Fei-Fei}.}
  \bibinfo{year}{2015}\natexlab{}.
\newblock \showarticletitle{{ImageNet Large Scale Visual Recognition
  Challenge}}.
\newblock \bibinfo{journal}{\emph{International Journal of Computer Vision}}
  (\bibinfo{year}{2015}).
\newblock


\bibitem[Saharia et~al\mbox{.}(2022)]%
        {saharia2022photorealistic}
\bibfield{author}{\bibinfo{person}{Chitwan Saharia}, \bibinfo{person}{William
  Chan}, \bibinfo{person}{Saurabh Saxena}, \bibinfo{person}{Lala Li},
  \bibinfo{person}{Jay Whang}, \bibinfo{person}{Emily Denton},
  \bibinfo{person}{Seyed Kamyar~Seyed Ghasemipour},
  \bibinfo{person}{Burcu~Karagol Ayan}, \bibinfo{person}{S~Sara Mahdavi},
  \bibinfo{person}{Rapha~Gontijo Lopes}, \bibinfo{person}{Tim Salimans},
  \bibinfo{person}{Jonathan Ho}, \bibinfo{person}{David~J Fleet}, {and}
  \bibinfo{person}{Mohammad Norouzi}.} \bibinfo{year}{2022}\natexlab{}.
\newblock \showarticletitle{{Photorealistic Text-to-Image Diffusion Models with
  Deep Language Understanding}}.
\newblock \bibinfo{journal}{\emph{arXiv preprint arXiv:2205.11487}}
  (\bibinfo{year}{2022}).
\newblock


\bibitem[Sarto et~al\mbox{.}(2023)]%
        {sarto2023positive}
\bibfield{author}{\bibinfo{person}{Sara Sarto}, \bibinfo{person}{Manuele
  Barraco}, \bibinfo{person}{Marcella Cornia}, \bibinfo{person}{Lorenzo
  Baraldi}, {and} \bibinfo{person}{Rita Cucchiara}.}
  \bibinfo{year}{2023}\natexlab{}.
\newblock \showarticletitle{{Positive-Augmented Contrastive Learning for Image
  and Video Captioning Evaluation}}. In \bibinfo{booktitle}{\emph{Proceedings
  of the IEEE/CVF Conference on Computer Vision and Pattern Recognition}}.
\newblock


\bibitem[Schramowski et~al\mbox{.}(2023)]%
        {schramowski2023safe}
\bibfield{author}{\bibinfo{person}{Patrick Schramowski},
  \bibinfo{person}{Manuel Brack}, \bibinfo{person}{Bj{\"o}rn Deiseroth}, {and}
  \bibinfo{person}{Kristian Kersting}.} \bibinfo{year}{2023}\natexlab{}.
\newblock \showarticletitle{{Safe Latent Diffusion: Mitigating Inappropriate
  Degeneration in Diffusion Models}}. In \bibinfo{booktitle}{\emph{Proceedings
  of the IEEE/CVF Conference on Computer Vision and Pattern Recognition}}.
\newblock


\bibitem[Schuhmann et~al\mbox{.}(2022)]%
        {schuhmann2022laion}
\bibfield{author}{\bibinfo{person}{Christoph Schuhmann},
  \bibinfo{person}{Romain Beaumont}, \bibinfo{person}{Richard Vencu},
  \bibinfo{person}{Cade Gordon}, \bibinfo{person}{Ross Wightman},
  \bibinfo{person}{Mehdi Cherti}, \bibinfo{person}{Theo Coombes},
  \bibinfo{person}{Aarush Katta}, \bibinfo{person}{Clayton Mullis},
  \bibinfo{person}{Mitchell Wortsman}, \bibinfo{person}{Patrick Schramowski},
  \bibinfo{person}{Srivatsa Kundurthy}, \bibinfo{person}{Katherine Crowson},
  \bibinfo{person}{Ludwig Schmidt}, \bibinfo{person}{Robert Kaczmarczyk}, {and}
  \bibinfo{person}{Jenia Jitsev}.} \bibinfo{year}{2022}\natexlab{}.
\newblock \showarticletitle{{LAION-5B: An open large-scale dataset for training
  next generation image-text models}}. In \bibinfo{booktitle}{\emph{Advances in
  Neural Information Processing Systems}}.
\newblock


\bibitem[Schuhmann et~al\mbox{.}(2021)]%
        {schuhmann2021laion}
\bibfield{author}{\bibinfo{person}{Christoph Schuhmann},
  \bibinfo{person}{Robert Kaczmarczyk}, \bibinfo{person}{Aran Komatsuzaki},
  \bibinfo{person}{Aarush Katta}, \bibinfo{person}{Richard Vencu},
  \bibinfo{person}{Romain Beaumont}, \bibinfo{person}{Jenia Jitsev},
  \bibinfo{person}{Theo Coombes}, {and} \bibinfo{person}{Clayton Mullis}.}
  \bibinfo{year}{2021}\natexlab{}.
\newblock \showarticletitle{{LAION-400M: Open Dataset of CLIP-Filtered 400
  Million Image-Text Pairs}}. In \bibinfo{booktitle}{\emph{Advances in Neural
  Information Processing Systems Workshops}}.
\newblock


\bibitem[Sha et~al\mbox{.}(2023)]%
        {sha2022fake}
\bibfield{author}{\bibinfo{person}{Zeyang Sha}, \bibinfo{person}{Zheng Li},
  \bibinfo{person}{Ning Yu}, {and} \bibinfo{person}{Yang Zhang}.}
  \bibinfo{year}{2023}\natexlab{}.
\newblock \showarticletitle{{DE-FAKE: Detection and Attribution of Fake Images
  Generated by Text-to-Image Diffusion Models}}. In
  \bibinfo{booktitle}{\emph{Proceedings of the ACM SIGSAC Conference on
  Computer and Communications Security}}.
\newblock


\bibitem[Vahdat and Kautz(2020)]%
        {vahdat2020nvae}
\bibfield{author}{\bibinfo{person}{Arash Vahdat} {and} \bibinfo{person}{Jan
  Kautz}.} \bibinfo{year}{2020}\natexlab{}.
\newblock \showarticletitle{{NVAE: A deep hierarchical variational
  autoencoder}}. In \bibinfo{booktitle}{\emph{Advances in Neural Information
  Processing Systems}}.
\newblock


\bibitem[Van~der Maaten and Hinton(2008)]%
        {van2008visualizing}
\bibfield{author}{\bibinfo{person}{Laurens Van~der Maaten} {and}
  \bibinfo{person}{Geoffrey Hinton}.} \bibinfo{year}{2008}\natexlab{}.
\newblock \showarticletitle{{Visualizing data using t-SNE}}.
\newblock \bibinfo{journal}{\emph{Journal of Machine Learning Research}}
  \bibinfo{volume}{9}, \bibinfo{number}{11} (\bibinfo{year}{2008}),
  \bibinfo{pages}{2579--2605}.
\newblock


\bibitem[Vaswani et~al\mbox{.}(2017)]%
        {vaswani2017attention}
\bibfield{author}{\bibinfo{person}{Ashish Vaswani}, \bibinfo{person}{Noam
  Shazeer}, \bibinfo{person}{Niki Parmar}, \bibinfo{person}{Jakob Uszkoreit},
  \bibinfo{person}{Llion Jones}, \bibinfo{person}{Aidan~N Gomez},
  \bibinfo{person}{{\L}ukasz Kaiser}, {and} \bibinfo{person}{Illia
  Polosukhin}.} \bibinfo{year}{2017}\natexlab{}.
\newblock \showarticletitle{Attention is all you need}. In
  \bibinfo{booktitle}{\emph{Advances in Neural Information Processing
  Systems}}.
\newblock


\bibitem[Verdoliva(2020)]%
        {verdoliva2020media}
\bibfield{author}{\bibinfo{person}{Luisa Verdoliva}.}
  \bibinfo{year}{2020}\natexlab{}.
\newblock \showarticletitle{Media forensics and deepfakes: an overview}.
\newblock \bibinfo{journal}{\emph{IEEE Journal of Selected Topics in Signal
  Processing}} \bibinfo{volume}{14}, \bibinfo{number}{5}
  (\bibinfo{year}{2020}), \bibinfo{pages}{910--932}.
\newblock


\bibitem[Wang et~al\mbox{.}(2020)]%
        {wang2020cnn}
\bibfield{author}{\bibinfo{person}{Sheng-Yu Wang}, \bibinfo{person}{Oliver
  Wang}, \bibinfo{person}{Richard Zhang}, \bibinfo{person}{Andrew Owens}, {and}
  \bibinfo{person}{Alexei~A Efros}.} \bibinfo{year}{2020}\natexlab{}.
\newblock \showarticletitle{{CNN-Generated Images Are Surprisingly Easy to
  Spot... for Now}}. In \bibinfo{booktitle}{\emph{Proceedings of the IEEE/CVF
  Conference on Computer Vision and Pattern Recognition}}.
\newblock


\bibitem[Wolter et~al\mbox{.}(2022)]%
        {wolter2022wavelet}
\bibfield{author}{\bibinfo{person}{Moritz Wolter}, \bibinfo{person}{Felix
  Blanke}, \bibinfo{person}{Raoul Heese}, {and} \bibinfo{person}{Jochen
  Garcke}.} \bibinfo{year}{2022}\natexlab{}.
\newblock \showarticletitle{Wavelet-packets for deepfake image analysis and
  detection}.
\newblock \bibinfo{journal}{\emph{Machine Learning}} (\bibinfo{year}{2022}),
  \bibinfo{pages}{1--33}.
\newblock


\bibitem[Wortsman et~al\mbox{.}(2022)]%
        {wortsman2022robust}
\bibfield{author}{\bibinfo{person}{Mitchell Wortsman}, \bibinfo{person}{Gabriel
  Ilharco}, \bibinfo{person}{Jong~Wook Kim}, \bibinfo{person}{Mike Li},
  \bibinfo{person}{Simon Kornblith}, \bibinfo{person}{Rebecca Roelofs},
  \bibinfo{person}{Raphael~Gontijo Lopes}, \bibinfo{person}{Hannaneh
  Hajishirzi}, \bibinfo{person}{Ali Farhadi}, \bibinfo{person}{Hongseok
  Namkoong}, {et~al\mbox{.}}} \bibinfo{year}{2022}\natexlab{}.
\newblock \showarticletitle{Robust fine-tuning of zero-shot models}. In
  \bibinfo{booktitle}{\emph{Proceedings of the IEEE/CVF Conference on Computer
  Vision and Pattern Recognition}}.
\newblock


\bibitem[Yang et~al\mbox{.}(2023)]%
        {yang2023avoid}
\bibfield{author}{\bibinfo{person}{Wenyuan Yang}, \bibinfo{person}{Xiaoyu
  Zhou}, \bibinfo{person}{Zhikai Chen}, \bibinfo{person}{Bofei Guo},
  \bibinfo{person}{Zhongjie Ba}, \bibinfo{person}{Zhihua Xia},
  \bibinfo{person}{Xiaochun Cao}, {and} \bibinfo{person}{Kui Ren}.}
  \bibinfo{year}{2023}\natexlab{}.
\newblock \showarticletitle{{AVoiD-DF: Audio-Visual Joint Learning for
  Detecting Deepfake}}.
\newblock \bibinfo{journal}{\emph{IEEE Transactions on Information Forensics
  and Security}}  \bibinfo{volume}{18} (\bibinfo{year}{2023}),
  \bibinfo{pages}{2015--2029}.
\newblock


\bibitem[Yu et~al\mbox{.}(2019)]%
        {yu2019attributing}
\bibfield{author}{\bibinfo{person}{Ning Yu}, \bibinfo{person}{Larry~S Davis},
  {and} \bibinfo{person}{Mario Fritz}.} \bibinfo{year}{2019}\natexlab{}.
\newblock \showarticletitle{{Attributing Fake Images to GANs: Learning and
  Analyzing GAN Fingerprints}}. In \bibinfo{booktitle}{\emph{Proceedings of the
  IEEE/CVF International Conference on Computer Vision}}.
\newblock


\bibitem[Zhang et~al\mbox{.}(2019)]%
        {zhang2019detecting}
\bibfield{author}{\bibinfo{person}{Xu Zhang}, \bibinfo{person}{Svebor Karaman},
  {and} \bibinfo{person}{Shih-Fu Chang}.} \bibinfo{year}{2019}\natexlab{}.
\newblock \showarticletitle{{Detecting and Simulating Artifacts in GAN Fake
  Images}}. In \bibinfo{booktitle}{\emph{Proceedings of the International
  Workshop on Information Forensics and Security}}.
\newblock


\bibitem[Zhu et~al\mbox{.}(1997)]%
        {zhu1997algorithm}
\bibfield{author}{\bibinfo{person}{Ciyou Zhu}, \bibinfo{person}{Richard~H
  Byrd}, \bibinfo{person}{Peihuang Lu}, {and} \bibinfo{person}{Jorge Nocedal}.}
  \bibinfo{year}{1997}\natexlab{}.
\newblock \showarticletitle{Algorithm 778: L-BFGS-B: Fortran subroutines for
  large-scale bound-constrained optimization}.
\newblock \bibinfo{journal}{\emph{ACM Trans. Math. Software}}
  \bibinfo{volume}{23}, \bibinfo{number}{4} (\bibinfo{year}{1997}),
  \bibinfo{pages}{550--560}.
\newblock


\bibitem[Zhu et~al\mbox{.}(2017)]%
        {zhu2017unpaired}
\bibfield{author}{\bibinfo{person}{Jun-Yan Zhu}, \bibinfo{person}{Taesung
  Park}, \bibinfo{person}{Phillip Isola}, {and} \bibinfo{person}{Alexei~A
  Efros}.} \bibinfo{year}{2017}\natexlab{}.
\newblock \showarticletitle{{Unpaired Image-To-Image Translation Using
  Cycle-Consistent Adversarial Networks}}. In
  \bibinfo{booktitle}{\emph{Proceedings of the IEEE/CVF International
  Conference on Computer Vision}}.
\newblock


\end{thebibliography}

\end{document}